\documentclass{article}

\PassOptionsToPackage{square,numbers}{natbib}
\usepackage[preprint]{neurips_2020}

\usepackage[T1]{fontenc}    
\usepackage{hyperref}       
\usepackage{url}            
\usepackage{booktabs}       
\usepackage{amsfonts}       
\usepackage{mathtools}
\usepackage{amsmath,amssymb}
\usepackage{amsthm,bm,framed,url}
\usepackage{nicefrac}       
\usepackage{microtype}      
\usepackage{graphicx}
\usepackage{subfigure}
\usepackage[ruled,vlined]{algorithm2e}
\usepackage{algpseudocode}
\usepackage{colortbl}
\usepackage{xcolor,colortbl}
\usepackage{threeparttable}
\usepackage{multicol}
\usepackage{multirow}
\usepackage{siunitx}
\usepackage{indentfirst,latexsym,bm}
\usepackage{mathrsfs}
\usepackage{enumitem}

\newcommand{\tabhead}[1]{\textbf{#1}}
\newcommand{\rothead}[1]{\rotatebox{30}{\tabhead{#1}}}

\SetKwInOut{Require}{Require}
\SetKwInOut{Init}{initialization}

\newcommand{\MomentumRate}{\ensuremath{\alpha}}

\newtheorem{definition}{Definition}
\newtheorem*{definition*}{Definition}
\newtheorem{theorem}{Theorem}
\newtheorem*{theorem*}{Theorem}
\newtheorem{remark}{Remark}
\newtheorem*{remark*}{Remark}

\newcommand{\rank}[1]{\text{rank}\{#1\}}
\newcommand{\trace}[1]{\text{Tr}\left\{#1\right\}}

\definecolor{mypink1}{rgb}{0.858, 0.188, 0.478}

\makeatletter
\def\hlinewd#1{%
  \noalign{\ifnum0=`}\fi\hrule \@height #1 \futurelet
   \reserved@a\@xhline}

\graphicspath{{figures/}}

\title{AdaS: Adaptive Scheduling of Stochastic Gradients}

\author{%
  Mahdi S.~Hosseini and Konstantinos N. Plataniotis \\
  University of Toronto, The Edward S. Rogers Sr. Department of Electrical \& Computer Engineering\\
  Toronto, Ontario, M5S 3G4, Canada \\
  \texttt{mahdi.hosseini@mail.utoronto.ca} \\
}

\begin{document}

\maketitle

\begin{abstract}
The choice of step-size used in Stochastic Gradient Descent (SGD) optimization is empirically selected in most training procedures. Moreover, the use of scheduled learning techniques such as Step-Decaying, Cyclical-Learning, and Warmup to tune the step-size requires extensive practical experience--offering limited insight into how the parameters update--and is not consistent across applications. This work attempts to answer a question of interest to both researchers and practitioners, namely \textit{``how much knowledge is gained in iterative training of deep neural networks?''} Answering this question introduces two useful metrics derived from the singular values of the low-rank factorization of convolution layers in deep neural networks. We introduce the notions of \textit{``knowledge gain''} and \textit{``mapping condition''} and propose a new algorithm called Adaptive Scheduling (AdaS) that utilizes these derived metrics to adapt the SGD learning rate proportionally to the rate of change in knowledge gain over successive iterations. Experimentation reveals that, using the derived metrics, AdaS exhibits: (a) faster convergence and superior generalization over existing adaptive learning methods; and (b) lack of dependence on a validation set to determine when to stop training. Code is available at \textcolor{mypink1}{\url{https://github.com/mahdihosseini/AdaS}}.
\end{abstract}

\section{Introduction}
Stochastic Gradient Descent (SGD), a first-order optimization method \cite{robbins1951stochastic, bottou2010large, bottou2012stochastic, schmidt2017minimizing}, has become the mainstream method for training over-parametrized models such as deep neural networks \cite{lecun2015deep, goodfellow2016deep}. Attempting to augment this method, SGD with momentum \cite{polyakmomentum1964, sutskever2013importance} accumulates the historically aligned gradients which helps in navigating past ravines and towards a more optimal solution. It eventually converges faster and exhibits better generalization compared to vanilla SGD. However, as the step-size (aka global learning rate) is mainly fixed for momentum SGD, it blindly follows these past gradients and can eventually overshoot an optimum and cause oscillatory behavior. From a practical standpoint (e.g. in the context of deep neural network \cite{bengio2012practical, schaul2013no, lecun2015deep}) it is even more concerning to deploy a fixed global learning rate as it often leads to poorer convergence, requires extensive tuning, and exhibits strong performance fluctuations over the selection range.

A handful of methods have been introduced over the past decade to solve the latter issues based on the adaptive gradient methods \cite{duchi2011, tieleman2012lecture, zeiler2012adadelta, kingma2014adam, dozat2016, SashankReddi2018on, Liu2020On, luo2018adaptive, gunes2018online, rolinek2018l4, vaswani2019painless, orabona2017training}. These methods can be represented in the general form:
\begin{equation}
        \Phi^k \longleftarrow \Phi^{k-1} - \frac{\eta_k}{\psi(g_1,\cdots,g_k)}\phi(g_1,\cdots,g_k),
\end{equation}
where for some $k$th iteration, $g_i$ is the stochastic gradient obtained at the $i$th iteration, $\phi(g_1,\cdots,g_k)$ is the gradient estimation, and $\eta_k/\psi(g_1,\cdots,g_k)$ is the adaptive learning rate, where $\psi(g_1,\cdots,g_k)$ generally relates to the square of the gradients. Each adaptive method therefore attempts to modify the gradient estimation (through the use of momentum) or the adaptive learning rate (through a difference choice of $\psi(g_1,\cdots,g_k)$). Furthermore, it is also common to subject $\eta_k$ to a manually set schedule for more optimal performance and better theoretical convergence guarantees.

Such methods were first introduced in \cite{duchi2011} (AdaGrad) by regulating the update size with the accumulated second order statistical measures of gradients which provides a robust framework for sparse gradient updates. The issue of vanishing learning rate caused by equally weighted accumulation of gradients is the main drawback of AdaGrad that was raised in \cite{tieleman2012lecture} (RMSProp), which utilizes the exponential decaying average of gradients instead of accumulation. A variant of first-order gradient measures was also introduced in \cite{zeiler2012adadelta} (AdaDelta), which solves the decaying learning rate problem using an accumulation window, providing a robust framework toward hyper-parameter tuning issues. The adaptive moment estimation in \cite{kingma2014adam} (AdaM) was introduced later to leverage both first and second moment measures of gradients for parameter updating. AdaM can be seen as the celebration of all three adaptive optimizers: AdaGrad, RMSProp and AdaDelta--solving the vanishing learning rate problem and offering a more optimal adaptive learning rate to improve in rapid convergence and generalization capabilities. Further improvements were made on AdaM using Nesterov Momentum \cite{dozat2016}, long-term memory of past gradients \cite{SashankReddi2018on}, rectified estimations \cite{Liu2020On}, dynamic bound of learning rate \cite{luo2018adaptive}, hyper-gradient descent method \cite{gunes2018online}, and loss-based step-size \cite{rolinek2018l4}. Methods based on line-search techniques \cite{vaswani2019painless} and coin betting \cite{orabona2017training} are also introduced to avoid bottlenecks caused by the hyper-parameter tuning issues in SGD.

The AdaM optimizer, as well as its other variants, has attracted many practitioners in deep learning for two main reasons: (1) it requires minimal hyper-parameter tuning effort; and (2) it provides an efficient convergence optimization framework. Despite the ease of implementation of such optimizers, there is a growing concern about their poor \textit{``generalization''} capabilities. They perform well on the given-samples i.e. training data (at times even better performance can be achieved compared to non-adaptive methods such as in \cite{loshchilov2016sgdr, smith2017cyclical, smith2019super}), but perform poorly on the out-of-samples i.e. test/evaluation data \cite{wilson2017marginal}. Despite various research efforts taken for adaptive learning methods, the non-adaptive SGD based optimizers (such as scheduled learning methods including Warmup Techniques \cite{loshchilov2016sgdr}, Cyclical-Learning \cite{smith2017cyclical, smith2019super}, and Step-Decaying \cite{goodfellow2016deep}) are still considered to be the golden frameworks to achieve better performance at the price of either more epochs for training and/or costly tuning for optimal hyper-parameter configurations given different datasets and models.

Our goal in this paper is twofold: (1) we address the above issues by proposing a new approach for adaptive methods in SGD optimization; and (2) we introduce new probing metrics that enable the monitoring and evaluation of quality of learning within layers of a Convolutional Neural Network (CNN). Unlike the general trend in most adaptive methods where the raw measurements from gradients are utilized to adapt the step-size and regulate the gradients (through different choices of adaptive learning rate or gradient estimation), we take a different approach and focus our efforts on the scheduling of the learning rate $\eta_k$ independently for each convolutional block. Specifically, we first ask \textit{``how much of the gradients are useful for SGD updates?''} and then translate this into a new concept we call the \textit{``knowledge gain''}, which is measured from the energy of low-rank factorization of convolution weights in deep layers. The knowledge gain defines the usefulness of gradients and adapts the next step-size $\eta_k$ for SGD updates. We summarize our contributions as follows:
\begin{enumerate}
\item The new concepts of \textit{``knowledge gain''} and \textit{``mapping condition''} are introduced to measure the quality of convolution weights that can be used in iterative training and to provide an answer to these questions: how well are the layers trained given a certain epoch steps? Is there enough information obtained via the sequence of updates?
\item We proposed a new adaptive scheduling algorithm for SGD called \textit{``AdaS''} which introduces minimal computational overhead over vanilla SGD and guarantees the increase of knowledge gain over consecutive epochs. AdaS adaptively schedules $\eta_k$ for every conv block and both generalizes well and outperforms previous adaptive methods e.g. AdaM. A pitching factor called \textit{gain-factor} is tuned in AdaS to trade off between fast convergence and greedy performance. Code is available at \textcolor{mypink1}{\url{https://github.com/mahdihosseini/AdaS}}.
\item Thorough experiments are conducted for image classification problems using various dataset and CNN models. We adopt different optimizers and compare their convergence speed and generalization characteristics to our AdaS optimizer.
\item A new probing tool based on knowledge gain and mapping condition is introduced to measure the quality of network training without requiring test/evaluation data. We investigate the relationship between our new quality metrics and performance results.
\end{enumerate}


\section{Knowledge Gain in CNN Training}\label{section_low_rank_factorization}
Central to our work is the notion of \textit{knowledge gain} measured from convolutional weights of CNNs. Consider the convolutional weights of a particular layer in a CNN defined by a four-way array (aka fourth-order tensor) ${\bf\Phi}\in\mathbb{R}^{{N_1}\times{N_2}\times{N_3}\times{N_4}}$, where $N_1$ and $N_2$ are the height and width of the convolutional kernels, and $N_3$ and $N_4$ correspond to the number of input and output channels, respectively. The feature mapping under this convolution operation follows ${\bf F}_{O}(:,:,\ell_4)=\sum^{N_3}_{\ell_3=1}{{\bf F}_{I}(:,:,\ell_3)\ast {\bf\Phi}(:,:,\ell_3,\ell_4)}$, where ${\bf F}_{I}$ and ${\bf F}_{O}$ are the input and output feature maps stacked in 3D volumes, and $\ell_4\in\{1,\ldots,N_4\}$ is the output index. The well-posedness of this feature mapping can be studied by the generalized spectral decomposition (i.e. SVD) form of the tensor arrays using the Tucker model \cite{kolda2009tensor, sidiropoulos2017tensor} in full-core tensor mode
\begin{equation}
{\bf\Phi} = \sum^{N_1}_{\ell_1=1}\sum^{N_2}_{\ell_2=1}\sum^{N_3}_{\ell_3=1}\sum^{N_4}_{\ell_4=1}
{\bf G}(\ell_1,\ell_2,\ell_3,\ell_4){\bf u}_{\ell_1} \circledcirc {\bf u}_{\ell_2} \circledcirc {\bf u}_{\ell_3} \circledcirc {\bf u}_{\ell_4},
\label{eq_lrf_1}
\end{equation}
where, the core ${\bf G}$ (containing singular values) is called a $(N_1,N_2,N_3,N_4)$-tensor, ${\bf u}_{\ell_i}\in\mathbb{R}^{N_i}$ is the factor basis for decomposition, and $\circledcirc$ is outer product operation. We use similar notations as in \cite{sidiropoulos2017tensor} for brevity. Note that ${\bf\Phi}$ can be at most of rank $(N_1,N_2,N_3,N_4)$. 

The tensor array in (\ref{eq_lrf_1}) is (usually) initialized by random noise sampling for CNN training such that the mapping under this tensor randomly spans the output dimensions (i.e. the diffusion of knowledge is fully random in the beginning with no learned structure). Throughout an iterative training framework, more knowledge is gained lying in the tensor space as a mixture of a low-rank manifold and perturbing noise. Therefore, it makes sense to decompose (factorize) the observing tensor within each layer of the CNN as ${\bf\Phi} = {\bf\hat{\Phi}} + {\bf E}$. This decomposes the observing tensor array into a low-rank tensor ${\bf\hat{\Phi}}$ containing the small-core tensor such that the error residues ${\bf E}=||{\bf\Phi}-{\bf\hat{\Phi}}||^2_{\text{F}}$ are minimized. A similar framework is also used in CNN compression \cite{lebedev2015speeding, Cheng_Tai_2016, kim2016compression, yu2017compressing}. A handful of techniques (e.g. CP/PRAFAC, TT, HT, truncated-MLSVD, Compression) can be found in \cite{kolda2009tensor, oseledets2011tensor, grasedyck2013literature, sidiropoulos2017tensor} to estimate such small-core tensor. The majority of these solutions are iterative and we therefore take a more careful consideration toward such low-rank decomposition.

An equivalent representation of the tensor decomposition (\ref{eq_lrf_1}) is the vector form ${\bf x}\triangleq \text{vec}\left({\bf\Phi}\right)=\left({\bf U}_1\otimes{\bf U}_2\otimes{\bf U}_3\otimes{\bf U}_4\right){\bf g}$, where $\text{vec}\left(\cdot\right)$ is a vector obtained by stacking all tensor elements column-wise, ${\bf g}=\text{vec}\left({\bf G}\right)$, $\otimes$ is the Kronecker product, and ${\bf U}_i$ is a factor matrix containing all bases ${\bf u}_{\ell_i}$ stacked in column form. Since we are interested in input and output channels of CNNs for decomposition, we use mode-3 and mode-4 vector expressions yielding two matrices
\begin{equation}
{\bf\Phi}_3=\left({\bf U}_1\otimes{\bf U}_2\otimes{\bf U}_4\right){\bf G}_3{\bf U}^T_3~~~\text{and}~~~{\bf\Phi}_4=\left({\bf U}_1\otimes{\bf U}_2\otimes{\bf U}_3\right){\bf G}_4{\bf U}^T_4,
\label{eq_lrf_2}
\end{equation}
where, ${\bf\Phi}_3\in\mathbb{R}^{N_1N_2N_4\times N_3}$, ${\bf\Phi}_4\in\mathbb{R}^{N_1N_2N_3\times N_4}$, and ${\bf G}_3$ and ${\bf G}_4$ are likewise reshaped forms of the core tensor ${\bf G}$. The tensor decomposition (\ref{eq_lrf_2}) is the equivalent representation to (\ref{eq_lrf_1}) decomposed at mode-3 and mode-4. Recall the matrix (two-way array) decomposition e.g. SVD such that ${\bf\Phi}_3={\bf U}{\bf \Sigma}{\bf V}^T$ where ${\bf U}\equiv{\bf U}_1\otimes{\bf U}_2\otimes{\bf U}_4$, ${\bf V}\equiv{\bf U}_3$, and ${\bf \Sigma}\equiv{\bf G}_3$ \cite{sidiropoulos2017tensor}. In other words, to decompose a tensor on a given mode, we first unfold the tensor (on the given mode) and then apply a decomposition method of interest such as SVD.

The presence of noise, however, is still a barrier for better understanding of the latter reshaped forms. Similar to \cite{lebedev2015speeding}, we revise our goal into low-rank matrix factorizations of ${\bf\Phi}_3={\bf\hat{\Phi}}_3+{\bf E}_3$ and ${\bf\Phi}_4={\bf\hat{\Phi}}_4+{\bf E}_4$, where a global analytical solution is given by the Variational Baysian Matrix Factorization (VBMF) technique in \cite{nakajima2013global} as a re-weighted SVD of the observation matrix. This method avoids unnecessary implementing an iterative algorithm.

Using the above decomposition framework, we introduce the following two definitions.
\begin{definition}{(Knowledge Gain).}\label{definition_knowledge_gain}
For convolutional weights in deep CNNs, define the knowledge gain across a particular channel (i.e. $d$-th dimension)
\begin{equation}
G_{d,p}({\bf{\Phi}})=\frac{1}{N_{d}\cdot\sigma^p_{1}({\bf\hat{\Phi}}_{d})}\sum^{N^{\prime}_{d}}_{i=1}{\sigma^p_{i}({\bf\hat{\Phi}}_{d})},
\label{eq_lrf_3}
\end{equation}
where, $\sigma_1\geq\sigma_2\geq\cdots\geq\sigma_{N^{\prime}_{d}}$ are the low-rank singular values of a single-channel convolutional weight in descending order, $N^{\prime}_{d}=\rank{{\bf\hat{\Phi}}_{d}}$, $d$ stands for dimension index, and $p\in\{1,2\}$.
\end{definition}
The notion of knowledge gain on the input tensor ${\bf{\Phi}}$ in (\ref{eq_lrf_3}) is in fact a direct measure of the norm energy of the factorized matrix.
\begin{remark}\label{remark_energy_norms}
Recall for $p=2$ that the summation of squared singular values from Definition \ref{definition_knowledge_gain} is equivalent to the {\em Frobenius} (norm) i.e. $\sum^{N^{\prime}_{d}}_{i=1}{\sigma^2_{i}({\bf\hat{\Phi}}_{d})}=\trace{{\bf\hat{\Phi}}^T_{d}{\bf\hat{\Phi}}_{d}}=||{\bf\hat{\Phi}}_{d}||^2_F$ \cite{horn2012matrix}. Also, for $p=1$ the summation of singular values is bounded between $||{\bf\hat{\Phi}}_{d}||_F\leq\sum^{N^{\prime}_{d}}_{i=1}{\sigma_{i}({\bf\hat{\Phi}}_{d})}\leq\sqrt{N^{\prime}_d}||{\bf\hat{\Phi}}_{d}||_F$.
\end{remark}
The energies here indicate the distance measure from the matrix separability obtained from low-rank factorization (similar to the index of inseparability in neurophysiology \cite{depireux2001spectro}). In other words, it measures the space span obtained by the low-rank structure. The division factors $N_{d}$ in (\ref{eq_lrf_3}) also normalize the gain $G_{p}\in[0,1]$ as a fraction of channel capacity. In this study we are mainly interested in third and fourth dimension measures (i.e. $d=\{3,4\}$).

\begin{definition}{(Mapping Condition).}\label{definition_mapping_condition}
For convolutional weights in deep CNNs, define the mapping condition across a particular channel (i.e. $d$-th dimension)
\begin{equation}
\kappa_{d}({\bf\Phi})={\sigma_{1}({\bf\hat{\Phi}}_{d})}/{\sigma_{N^{\prime}_{d}}({\bf\hat{\Phi}}_{d})},
\label{eq_lrf_4}
\end{equation}
where, $\sigma_{1}$ and $\sigma_{N^{\prime}_d}$ are the maximum and minimum low-rank singular values of a single-channel convolutional weight, respectively.
\end{definition}
Recall the matrix-vector calculation form by mapping the input vector into the output vector, where its numerical stability is defined by the matrix condition number as a relative ratio of maximum to minimum singular values \cite{horn2012matrix}. The convolution operations in CNNs follow a similar concept by mapping input feature images into output features. Accordingly, the mapping condition of the convolutional layers in CNN is defined by (\ref{eq_lrf_4}) as a direct measurement of condition number of low-rank factorized matrices: indicating the well-posedness of  convolution operation.

\section{Adapting Stochastic Gradient Descent with Knowledge Gain}\label{sec_AdaS}
As an optimization method in deep learning, SGD typically attempts to minimize the loss functions of large networks \cite{bottou2010large, bottou2012stochastic, lecun2015deep, goodfellow2016deep}. Consider the updates on the convolutional weights ${\bf\Phi}^{k}$ using this optimization
\begin{equation}
{\bf\Phi}^{k}\longleftarrow{\bf\Phi}^{k-1} - \eta_k\nabla\tilde{f}_k({\bf\Phi}^{k-1})~~~\text{for}~~~k\in\{(t-1){K}+1,\cdots,t{K}\},
\label{eq_AdaS_1}
\end{equation}
where $t$ and $K$ correspond to epoch index and number of mini-batches, respectively, $\nabla\tilde{f}_k({\bf\Phi}^{k-1})=1/|\Omega_k|\sum_{i\in\Omega_k}{\nabla f_i({\bf\Phi}^{k-1})}$ is the average stochastic gradients on $k$th mini-batch that are randomly selected from a batch of $n$-samples $\Omega_k\subset\{1,\cdots,n\}$, and $\eta_k$ defines the step-size taken toward the opposite direction of average gradients. The selection of step-size $\eta_k$ can be either adaptive with respect to the statistical measure from gradients \cite{duchi2011, zeiler2012adadelta, tieleman2012lecture, kingma2014adam, dozat2016, SashankReddi2018on, luo2018adaptive, vaswani2019painless, Liu2020On} or could be subject to change in different scheduled learning regimes \cite{loshchilov2016sgdr, smith2017cyclical, smith2019super, loshchilov2016sgdr}.

In the scheduled learning rate method, the step-size is usually fixed for every $t$th epoch (i.e. for all $K$ mini-batch updates) and changes according to the schedule assignment for the next epoch (i.e. $\eta_k\equiv\eta(t)$). We setup our problem by accumulating all observed gradients throughout $K$ mini-batch updates within the $t$th epoch
\begin{equation}
{\bf\Phi}^{k_b}={\bf\Phi}^{k_a} - \eta(t)\sum^{k_b}_{k=k_a}\nabla\tilde{f}_k({\bf\Phi}^{k}),
\label{eq_AdaS_2}
\end{equation}
where $k_a=(t-1){K}+1$ and $k_b=t{K}$. Note that the significance of updates in (\ref{eq_AdaS_2}) from $k_a$th to $k_b$th iteration is controlled by the step-size $\eta(t)$, which directly impacts the rate of the knowledge gain.

Here we provide satisfying conditions on the step-size for increasing the knowledge gain in SGD. 
\begin{theorem}{(Increasing Knowledge Gain for SGD).}\label{theorem_knowledge_gain_SGD}
Using the knowledge gain from Definition \ref{eq_lrf_3} and setting the step-size of Stochastic Gradient Descent (SGD) proportionate to
\begin{equation}
\eta = \zeta\left[{G({\bf\Phi}^{k_b})} - {G({\bf\Phi}^{k_a})}\right]
\label{eq_AdaS_3}
\end{equation}
will guarantee the monotonic increase of the knowledge gain i.e. $G({\bf\Phi}^{k_b})\geq G({\bf\Phi}^{k_a})$ for some existing lower bound $\eta\geq{\eta_0}$ and $\zeta\geq{0}$.
\end{theorem}
The proof of Theorem \ref{theorem_knowledge_gain_SGD} is provided in the Appendix-A. The step-size in (\ref{eq_AdaS_3}) is proportional to the knowledge gain through the updating scheme in SGD where we update the value in every epoch. Therefore, the computational overhead on vanilla SGD is only limited to calculating the knowledge gain for each convolutional layer in the CNN for every epoch. This overhead is minimal due to the empirical solution provided by the low-rank factorization method (EVBMF) in \cite{nakajima2013global}.

\section{AdaS Algorithm}\label{sec_adas_algorithm}
We formulate the update rule for AdaS using SGD with Momentum as follows
\begin{align}
        \eta(t, \ell) &\leftarrow \beta\cdot\eta(t-1, \ell) + \zeta \cdot[\bar{G}(t,\ell)-\bar{G}(t-1,\ell)]\\
        v^{k}_{\ell} &\leftarrow \MomentumRate\cdot v^{k-1}_{\ell} - \eta(t,\ell)\cdot{g^{k}_{\ell}}\\
        \theta^k_{\ell} &\leftarrow \theta^{k-1}_{\ell} + v_{\ell}^k
\end{align}
where $k$ is the current mini-batch, $t$ is the current epoch iteration, $\ell$ is the conv block index, $\bar{G}(\cdot)$ is the average knowledge gain obtained from both mode-3 and mode-4 decompositions, $v$ is the velocity term, and $\theta$ are the learnable parameters.

The pseudo-code for our proposed algorithm \textit{AdaS} is presented in Algorithm \ref{algorithm_adas}. Each convolution block in the CNN is assigned an index $\{\ell\}^{L}_{\ell=1}$ where all learnable parameters (e.g. conv, biases, batch-norms, etc) are called using this index. The goal in AdaS is firstly to callback the convolutional weights within each block, secondly to apply low-rank matrix factorization on the unfolded tensors, and finally approximate the overall knowledge gain $G^t_{\ell}$ and mapping condition $\kappa^t_{\ell}$. The approximation is done once every epoch and introduces minimal computational overhead over the rest of the optimization framework. The learning rate is computed relative to the rate of change in knowledge gain over two consecutive epoch updates (from previous to current). The learning rate $\eta(t,\ell)$ is then further updated by an exponential moving average called the gain-factor, with hyper-parameter $\beta$, to accumulate the history of knowledge gain over the sequence of epochs. In effect, $\beta$ controls the tradeoff between convergence speed and training accuracy of AdaS. An ablative study on the effect of this parameter is provided in the Appendix-B. The computed step-sizes for all conv-blocks are then passed through the SGD optimization framework for adaptation. Note that the same step-size is used within each block for all learnable parameters. Code is available at \textcolor{mypink1}{\url{https://github.com/mahdihosseini/AdaS}}.

\begin{algorithm}[H]
\SetAlgoLined
\Require{batch size $n$, \# of epochs $T$, \# of conv blocks $L$, initial step-sizes $\left\{\eta(0, \ell)\right\}^{L}_{\ell=1}$, initial momentum vectors $\left\{v_{\ell}^0\right\}_{{\ell}=1}^L$, initial parameter vectors $\left\{\theta_{\ell}^0\right\}_{\ell=1}^L$, SGD momentum rate $\MomentumRate \in [0, 1)$, AdaS gain factor $\beta\in [0, 1)$, knowledge gain hyper-parameter $\zeta=1$, minimum learning rate $\eta_{\min} > 0$}
\For{$t$ = $1$ : $T$}{
\For{$\ell$ = $1$ : $L$}{
        1. unfold tensors using (\ref{eq_lrf_2}): ${\bf\Phi}_3\leftarrow\text{mode-3}\left({\bf\Phi}^{t}_{\ell}\right)$ and ${\bf\Phi}_4\leftarrow\text{mode-4}\left({\bf\Phi}^{t}_{\ell}\right)$ \\
2. apply low-rank factorization \cite{nakajima2013global}: ${\bf\hat{\Phi}}_3\leftarrow \texttt{EVBMF}\left({\bf\Phi}_3\right)$ and ${\bf\hat{\Phi}}_4\leftarrow \texttt{EVBMF}\left({\bf\Phi}_4\right)$\\
3. compute average knowledge gain using (\ref{eq_lrf_3}): $\overline{G}(t,\ell)\leftarrow [G_{3,1}({\bf\Phi}) + G_{4,1}({\bf\Phi})]/2$\\
4. compute average mapping condition using (\ref{eq_lrf_4}): $\kappa(t,\ell)\leftarrow [\kappa_{3}({\bf{\Phi}}) + \kappa_{4}({\bf{\Phi}})]/2$\\
5. compute step momentum: $\eta(t,\ell)\leftarrow \beta\cdot\eta(t-1,\ell) + \zeta\cdot{[\overline{G}(t,\ell) - \overline{G}(t-1,\ell)]}$\\
6. lower bound the learning rate: $\eta(t,\ell)\leftarrow \max{(\eta(t,\ell), \eta_{\min})}$
}
randomly shuffle dataset, generate $K$ mini-batches \(\{\Omega_k\subset\{1,\cdots,n\}\}^K_{k=1}\)\\
\For{$k = (t-1){K}+1 : tK$}{
1. compute gradient: $g^{k}_{\ell} \leftarrow \frac{1}{|\Omega_k|}{\sum\limits_{i\in\Omega_k}}\nabla_\Phi f((x^{(i)}, y^{(i)});\ \Phi^{k-1}_{\ell}),~~~\ell\in\{1,\cdots,L\}$\\
2. compute the velocity term: $v^{k}_{\ell} \leftarrow \MomentumRate\cdot v^{k-1}_{\ell} - \eta(t,\ell)\cdot{g^{k}_{\ell}}$\\
3. apply update: $\theta^{k}_{\ell} \leftarrow \theta^{k-1}_{\ell} + v^{k}_{\ell}$
}}
\caption{Adaptive Scheduling (AdaS) for SGD with Momentum}
\label{algorithm_adas}
\end{algorithm}
\section{Experiments}\label{sec_experiments}
We compare our AdaS algorithm to several adaptive and non-adaptive optimizers in the context of image classification. In particular, we implement AdaS with SGD with momentum, four adaptive methods i.e. AdaGrad \cite{duchi2011}, RMSProp \cite{goodfellow2016deep}, AdaM \cite{kingma2014adam}, AdaBound \cite{luo2018adaptive}, and two non-adaptive momentum SGDs guided by scheduled learning techniques i.e. OneCyleLR (also known as the super-convergence method) \cite{smith2019super} and SGD with StepLR (step decaying) \cite{goodfellow2016deep}. We further investigate the dependencies of CNN training quality to knowledge gain and mapping conditions and provide useful insights into the usefulness of different optimizers for training different models. For details on ablative studies and a complete set of experiments, please refer to the Appendix-B.

\subsection{Experimental Setup}
We investigate the efficacy of AdaS with respect to variations in the number of deep layers using VGG16 \cite{simonyan2014very} and ResNet34 \cite{he2016deep} and the number of classes using the standard CIFAR-10 and CIFAR-100 datasets \cite{krizhevsky2009learning} for training. The details of pre-processing steps, network implementation and training/testing frameworks are adopted from the CIFAR GitHub repository\footnote{\url{https://github.com/kuangliu/pytorch-cifar}} using PyTorch. We set the initial learning rates of AdaGrad, RMSProp and AdaBound to $\eta_0=\{1\text{e-}2, 3\text{e-}4, 1\text{e-}3\}$ per their suggested default values. We further followed the suggested tuning in \cite{wilson2017marginal} for AdaM ($\eta_0=3\text{e-}4$ for VGG16 and $\eta_0=1\text{e-}3$ for ResNet34) and SGD-StepLR ($\eta_0=1\text{e-}1$ dropping half magnitude every $25$ epochs) to achieve the best performance. For SGD-1CycleLR we set $50$ epochs for the whole cycle and found the best configuration ($\eta_0=3\text{e-}2$ for VGG16 and $\eta_0=4\text{e-}2$ for ResNet34). To configure the best initial learning rate for AdaS, we performed a dense grid search and found the values for VGG16 and ResNet34 to be $\eta_0=\{5\text{e-}3, 3\text{e-}2\}$. Despite the differences in optimal values that are independently obtained for each network, the optimizer performance is fairly robust relative to changes in these values. Each model is trained for $250$ epochs in $5$ independent runs and average test accuracy and training losses are reported. The mini-batch size is also set to $|\Omega_k|=128$.

\begin{figure}[htp]
	\centerline{
		\subfigure[\tiny{Test Acc--VGG16--CIFAR10}]{\includegraphics[width=.24\textwidth]{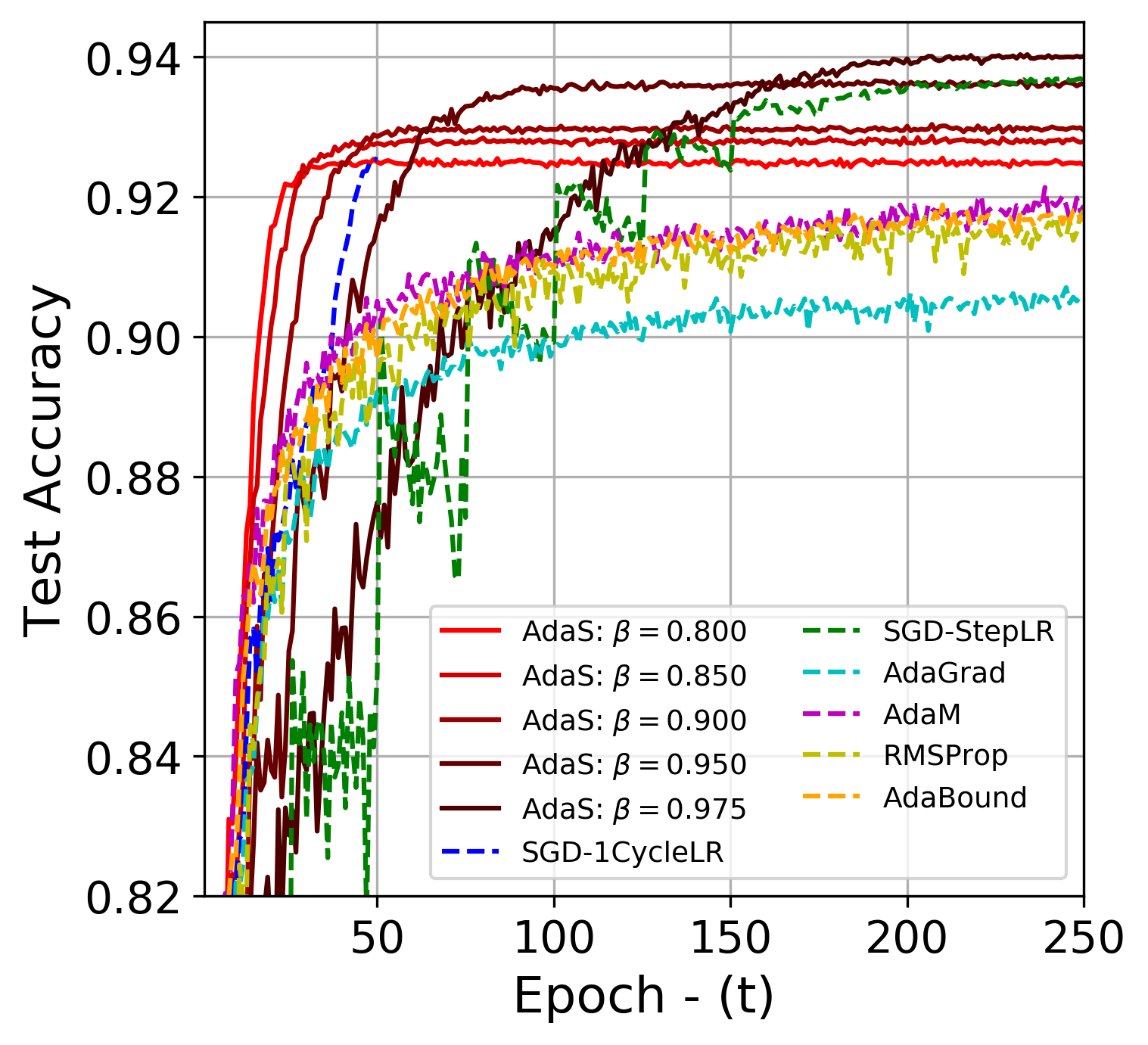}}
		\subfigure[\tiny{Test Acc--ResNet34--CIFAR10}]{\includegraphics[width=.24\textwidth]{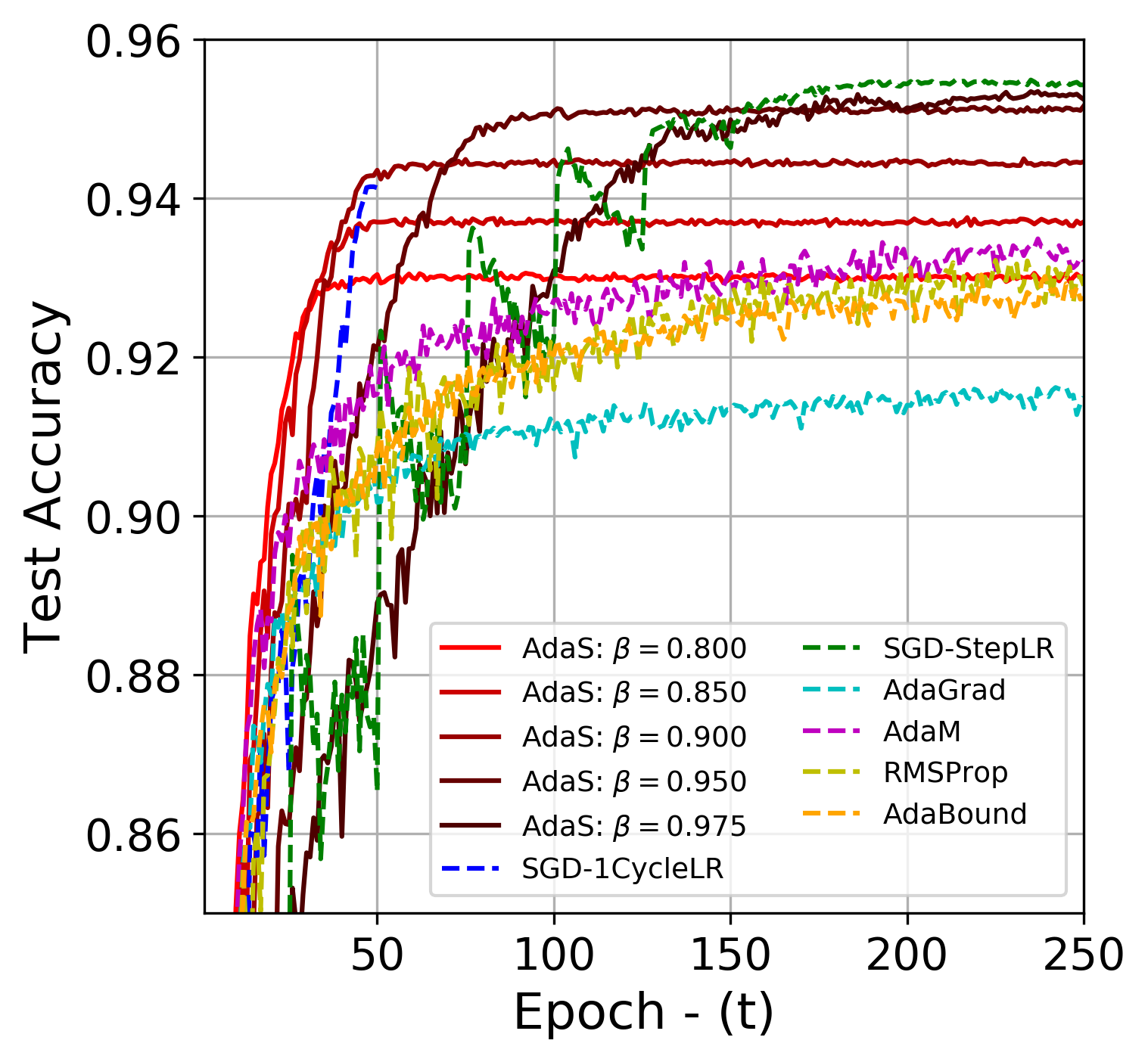}}
		\subfigure[\tiny{Test Acc--VGG16--CIFAR100}]{\includegraphics[width=.24\textwidth]{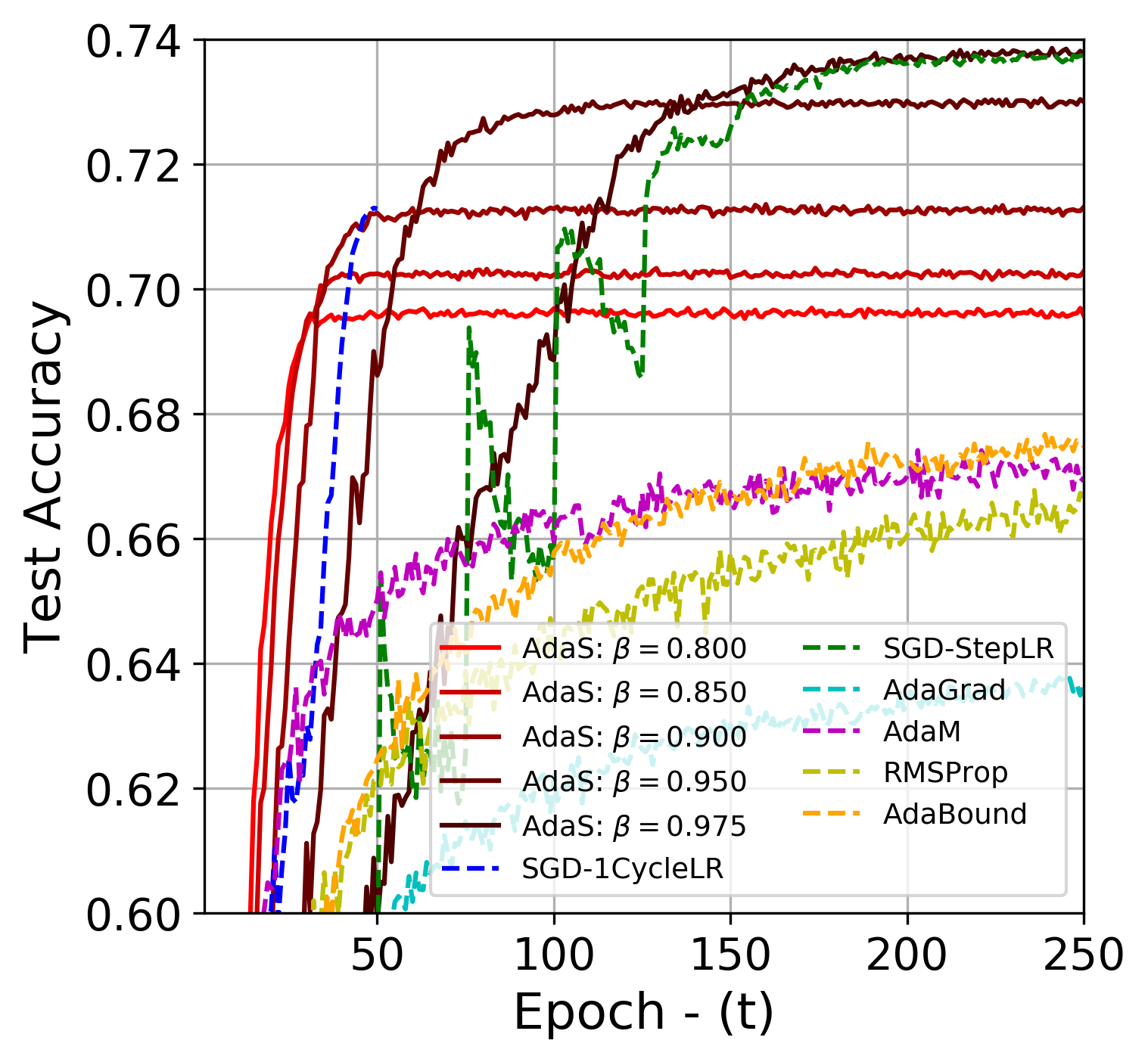}}
		\subfigure[\tiny{Test Acc--ResNet34--CIFAR100}]{\includegraphics[width=.24\textwidth]{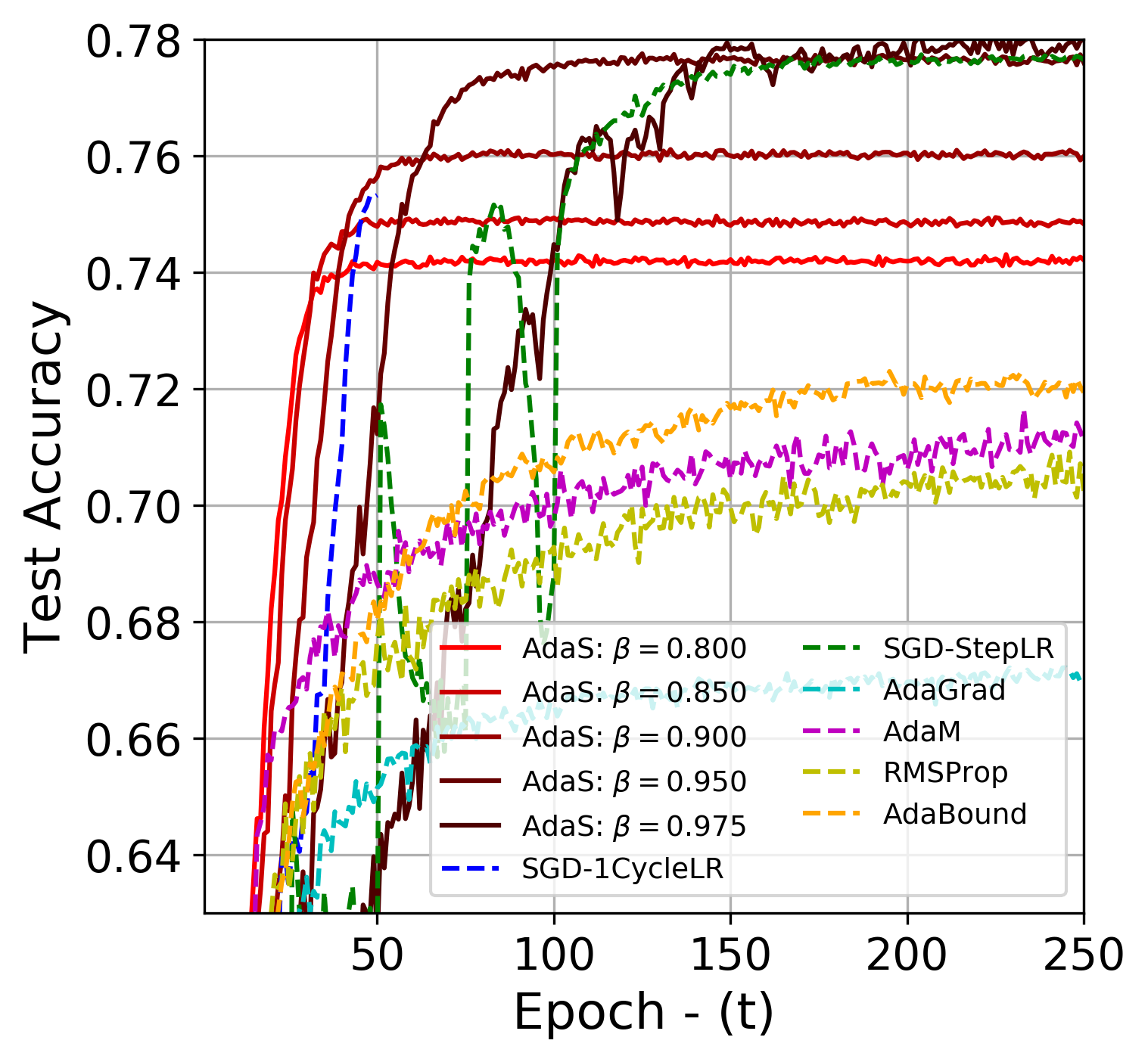}}
	}
	\centerline{
		\subfigure[\tiny{Train Loss--VGG16--CIFAR10}]{\includegraphics[width=.24\textwidth]{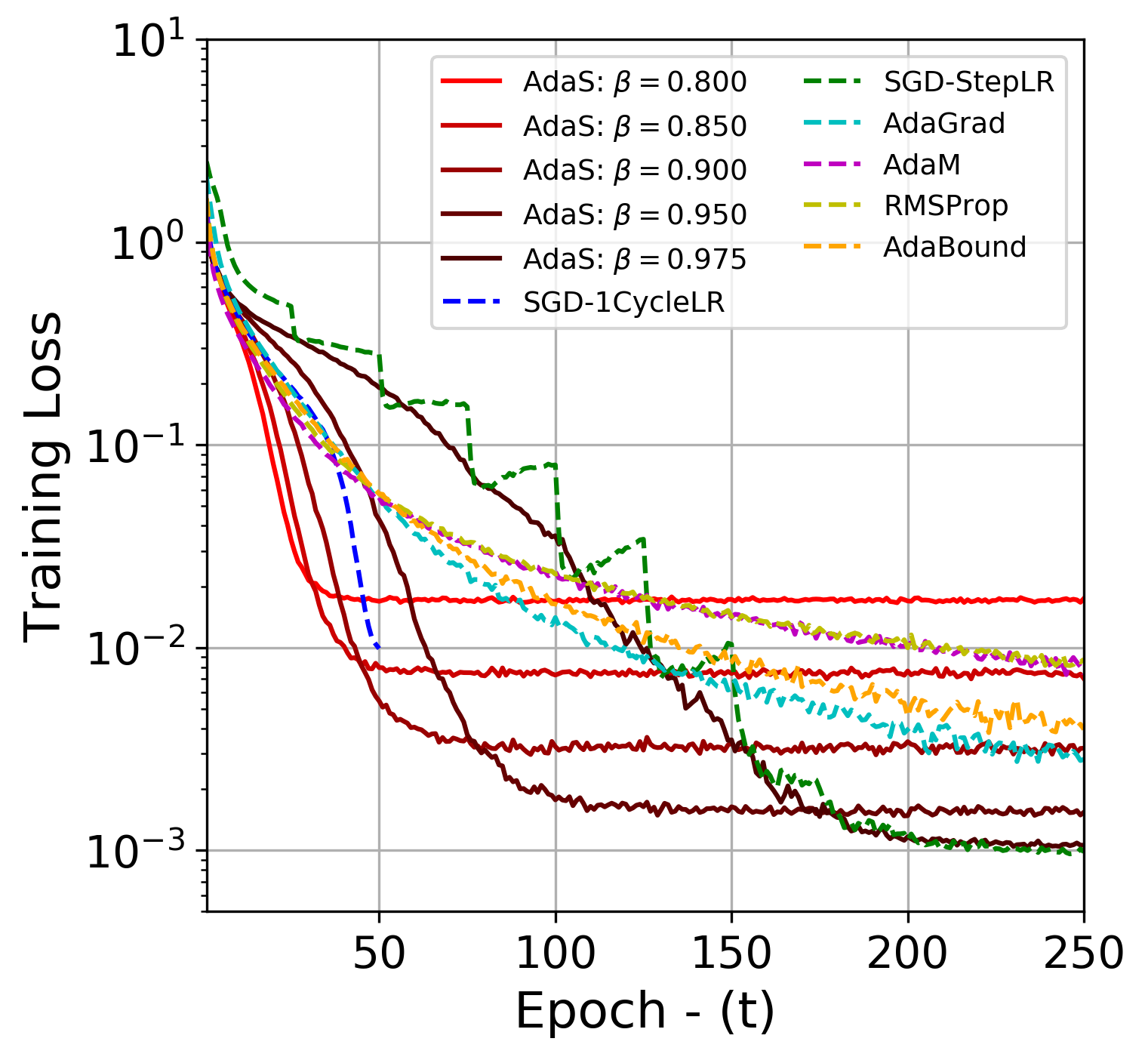}}
		\subfigure[\tiny{Train Loss--ResNet34--CIFAR10}]{\includegraphics[width=.24\textwidth]{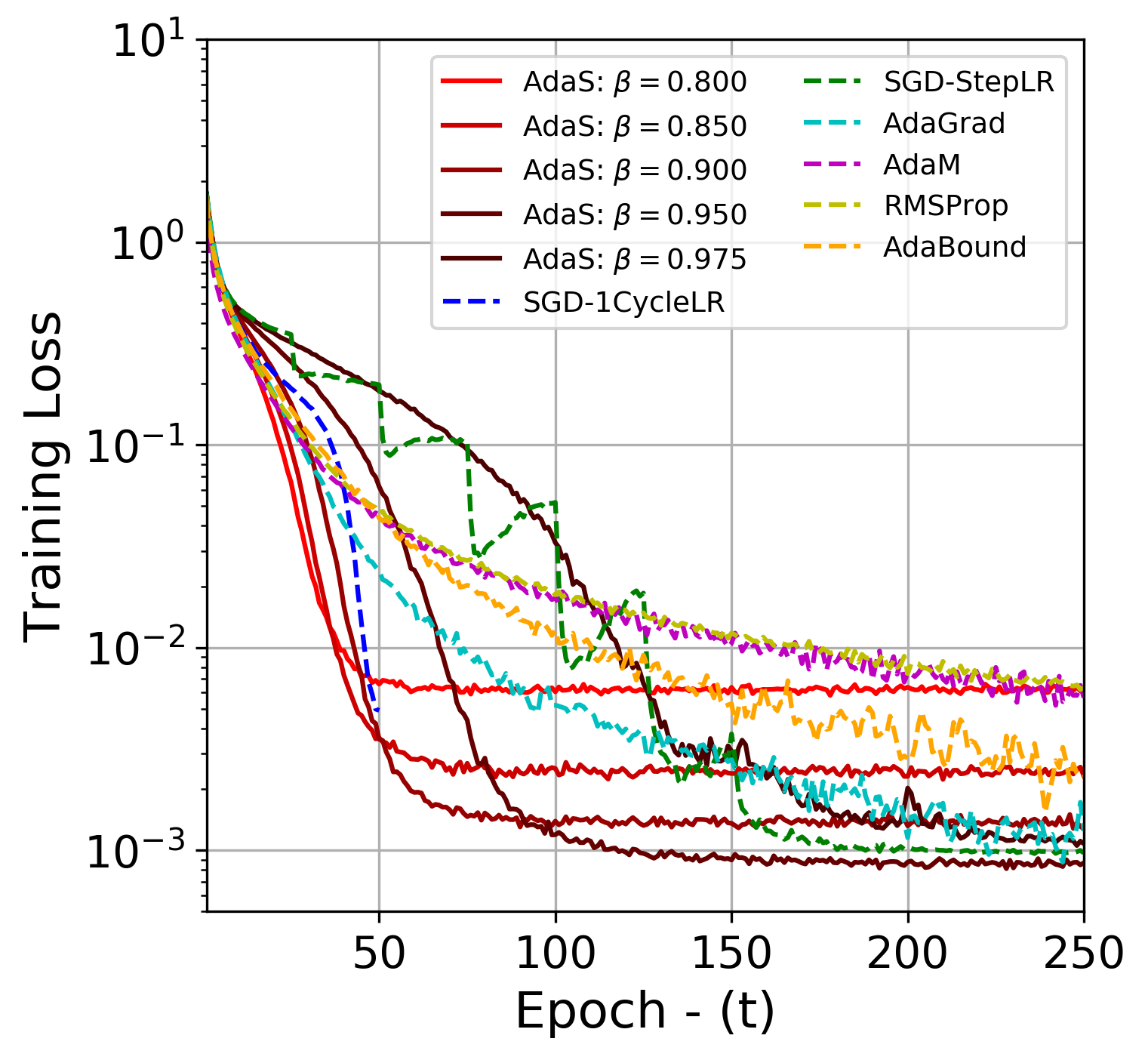}}
		\subfigure[\tiny{Train Loss--VGG16--CIFAR100}]{\includegraphics[width=.24\textwidth]{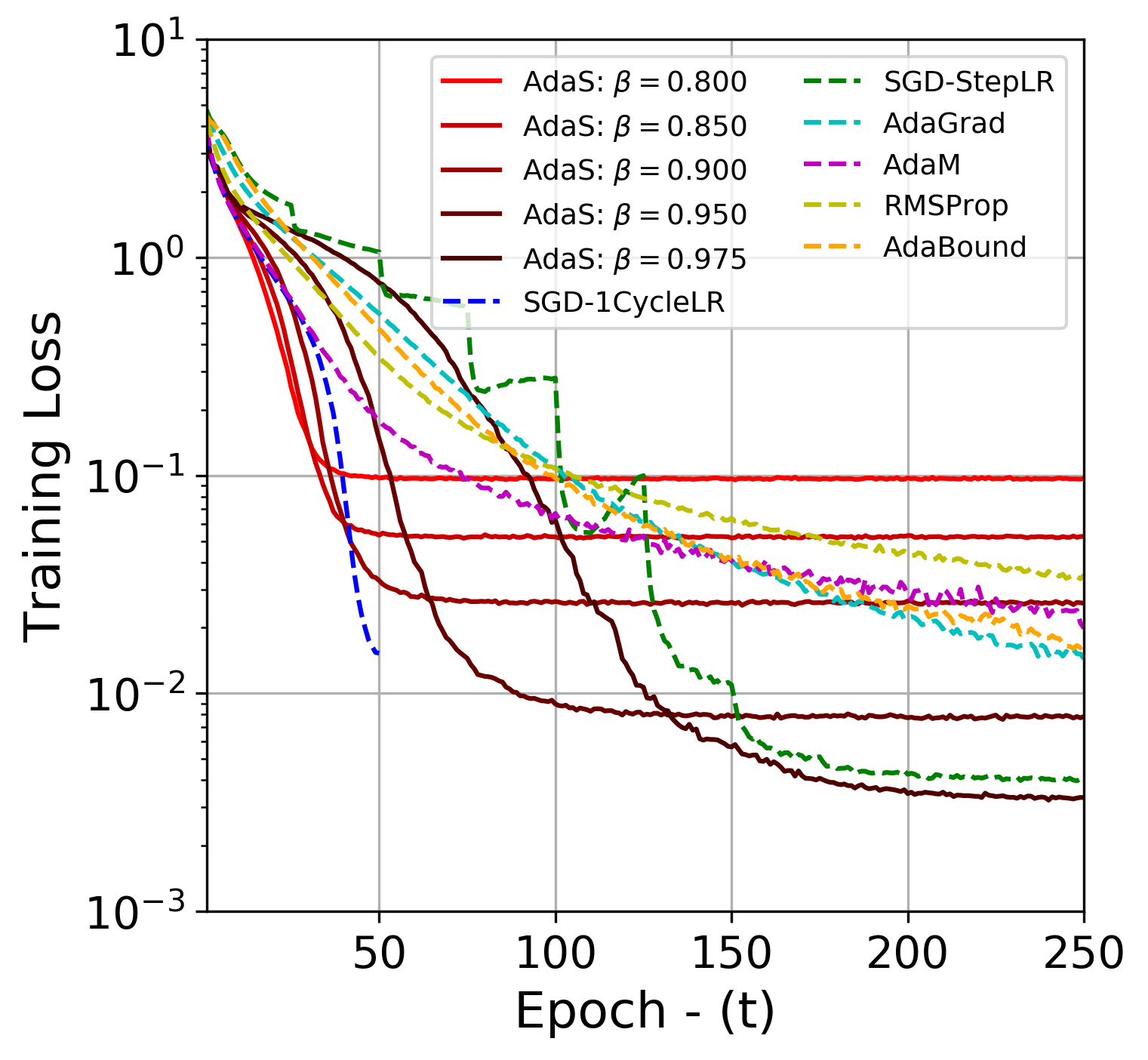}}
		\subfigure[\tiny{Train Loss--ResNet34--CIFAR100}]{\includegraphics[width=.24\textwidth]{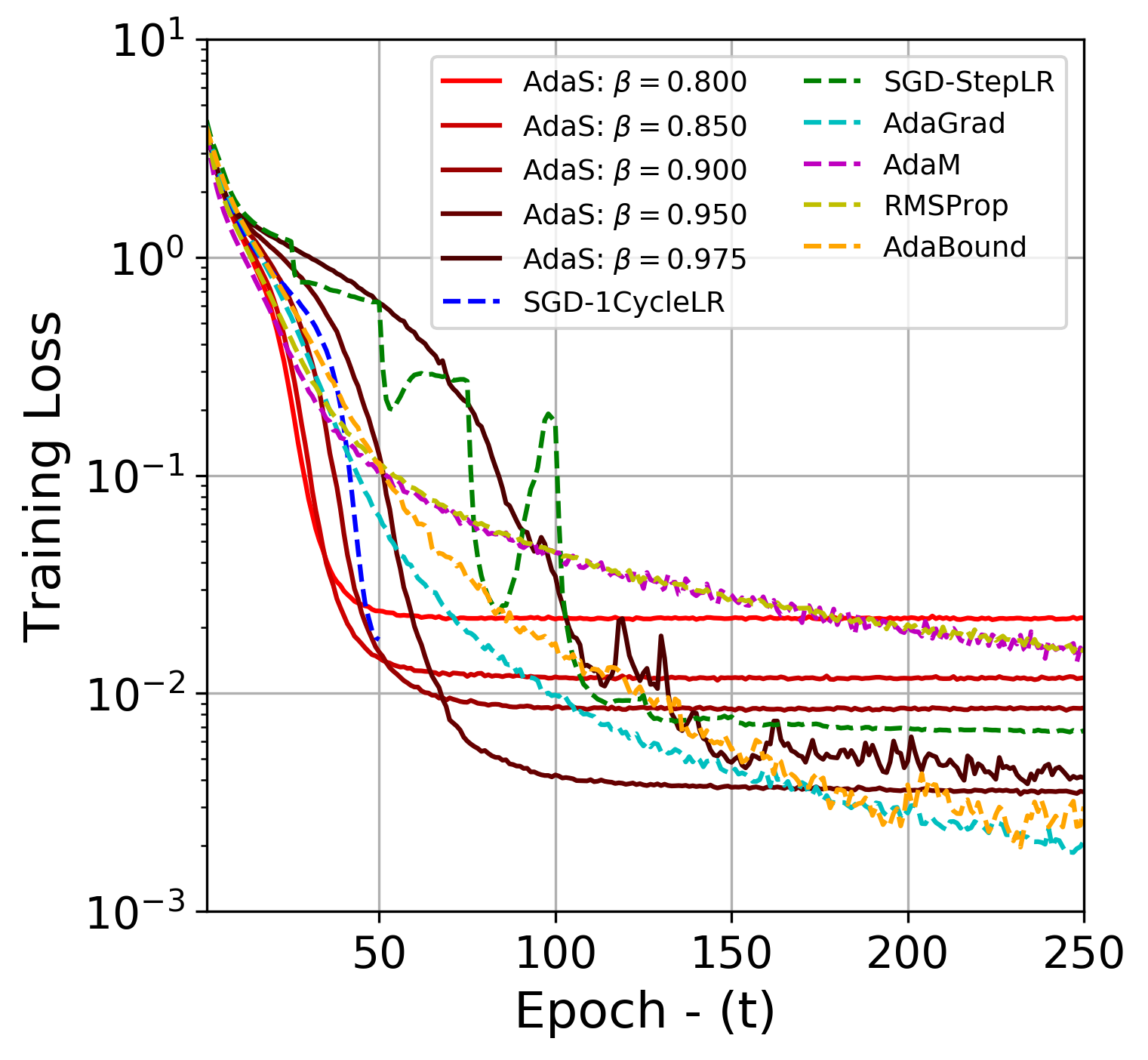}}
	}
	\caption{Training performance using different optimizers across two different datasets (i.e. CIFAR10 and CIFAR100) and two different CNNs (i.e. VGG16 and ResNet34)}
	\label{fig_comparison_all_methods}
\end{figure}

\subsection{Image Classification Problem}
We first empirically evaluate the effect of the gain-factor $\beta$ on AdaS convergence by defining eight different grid values (i.e. $\beta\in\{0.8, 0.825,\cdots,0.975\}$). The trade-off between the selection of different values of $\beta$ is demonstrated in Figure \ref{fig_comparison_all_methods} (complete ablation study is provided in Appendix-B). Here, lower $\beta$ translates to faster convergence, whereas setting it to higher values yields better final performance--at the cost of requiring more epochs for training. The performance comparison of optimizers is also overlaid in the same figure, where AdaS (with lower $\beta$) surpasses all adaptive and non-adaptive methods by a large margin in both test accuracy and training loss during the initial stages of training (i.e. $\text{epoch}<50$), where as SGD-StepLR and AdaS (with higher $\beta$) eventually overtake the other methods with more training epochs. Furthermore, AdaGrad, RMSProp, AdaM, and AdaBound all achieve similar or sometimes even lower training losses compared to AdaS (including the two non-adaptive methods), but attain lower test accuracies. Such controversial results were also reported in \cite{wilson2017marginal} where adaptive optimizers generalize worse compared to non-adaptive methods. In retrospect, we claim here that AdaS solves this issue by generalizing better than other adaptive optimizers.

We further provide quantitative results on the convergence of all optimizers trained on ResNet34 in Table \ref{tab_convergence_test_accuracy} with a fixed number of training epochs. The rank consistency of AdaS (using two different gain factors of low $\beta=0.85$ and high $\beta=0.95$ values) over other optimizers is evident. For instance, AdaS$^{\beta=0.850}$ gains $3.63\%$ test accuracy (with half confidence interval) over the second best optimizer AdaM on CIFAR-100 trained with $25$ epochs.

\begin{table}[htp]
  \tiny
  \caption{Image classification performance (test accuracy) of ResNet34 on CIFAR-10 and CIFAR-100 with fixed budget epochs. Four adaptive (AdaGrad, RMSProp, AdaM, AdaS) and one non-adaptive (SGD-StepLR) optimizers are deployed for comparison.}
  \label{tab_convergence_test_accuracy}
  \centering  
  \begin{threeparttable}
    \tiny
    \begin{tabular*}{\textwidth}
      {p{0.1cm}p{0.2cm}p{1.7cm}p{1.7cm}p{1.7cm}p{1.7cm}p{1.7cm}p{1.7cm}}    
      \toprule      
      & {Epoch} & {AdaGrad} & {RMSProp} & {AdaM} & {SGD-StepLR} & {AdaS$^{\beta=0.850}$} & {AdaS$^{\beta=0.950}$}\\
			\toprule
			\multirow{7}{*}{\rotatebox[origin=c]{90}{CIFAR-10}} 
			& $25$ & $0.8859\pm{0.47\%}$ & $0.8916\pm{0.71\%}$ & ${\bf 0.8957\pm{0.73\%}}$ & $0.8325\pm{2.79\%}$ & ${\bf 0.9136\pm{0.23\%}}$ & $0.8611\pm{1.67\%}$\\
			& $50$ & $0.9017\pm{0.61\%}$ & $0.9086\pm{0.58\%}$ & $0.9154\pm{0.35\%}$ & $0.8653\pm{1.67\%}$ & ${\bf 0.9370\pm{0.13\%}}$ & ${\bf 0.9209\pm{0.52\%}}$\\
			& $75$ & $0.9103\pm{0.18\%}$ & $0.9139\pm{0.78\%}$ & $0.9211\pm{0.26\%}$ & $0.9067\pm{0.38\%}$ & ${\bf 0.9372\pm{0.14\%}}$ & ${\bf 0.9472\pm{0.20\%}}$\\
			& $100$ & $0.9109\pm{0.25\%}$ & $0.9159\pm{0.77\%}$ & $0.9271\pm{0.27\%}$ & $0.9225\pm{0.29\%}$ & ${\bf 0.9372\pm{0.12\%}}$ & ${\bf 0.9510\pm{0.18\%}}$\\
			& $150$ & $0.9134\pm{0.37\%}$ & $0.9269\pm{0.33\%}$ & $0.9307\pm{0.33\%}$ & ${\bf 0.9464\pm{0.09\%}}$ & $0.9370\pm{0.13\%}$ & ${\bf 0.9510\pm{0.11\%}}$\\
			& $200$ & $0.9140\pm{0.15\%}$ & $0.9287\pm{0.30\%}$ & $0.9317\pm{0.30\%}$ & ${\bf 0.9544\pm{0.10\%}}$ & $0.9368\pm{0.11\%}$ & ${\bf 0.9508\pm{0.15\%}}$\\
			& $250$ & $0.9149\pm{0.24\%}$ & $0.9290\pm{0.29\%}$ & $0.9322\pm{0.36\%}$ & ${\bf 0.9543\pm{0.08\%}}$ & $0.9370\pm{0.17\%}$ & ${\bf 0.9516\pm{0.12\%}}$\\
			\toprule
			\multirow{7}{*}{\rotatebox[origin=c]{90}{CIFAR-100}} 
			& $25$ & $0.6221\pm{0.70\%}$ & $0.6341\pm{1.14\%}$ & ${\bf 0.6653\pm{0.46\%}}$ & $0.5545\pm{1.45\%}$ & ${\bf 0.7016\pm{0.27\%}}$ & $0.5981\pm{1.55\%}$\\
			& $50$ & $0.6515\pm{0.27\%}$ & $0.6769\pm{0.62\%}$ & $0.6866\pm{0.46\%}$ & $0.6217\pm{1.68\%}$ & ${\bf 0.7479\pm{0.23\%}}$ & ${\bf 0.7123\pm{0.51\%}}$\\
			& $75$ & $0.6618\pm{0.38\%}$ & $0.6837\pm{0.50\%}$ & $0.6975\pm{0.49\%}$ & $0.6611\pm{1.79\%}$ & ${\bf 0.7491\pm{0.26\%}}$ & ${\bf 0.7714\pm{0.30\%}}$\\
			& $100$ & $0.6658\pm{0.38\%}$ & $0.6928\pm{0.29\%}$ & $0.6978\pm{0.27\%}$ & $0.6878\pm{0.97\%}$ & ${\bf 0.7494\pm{0.26\%}}$ & ${\bf 0.7752\pm{0.31\%}}$\\
			& $150$ & $0.6691\pm{0.31\%}$ & $0.6996\pm{0.48\%}$ & $0.7045\pm{0.42\%}$ & ${\bf 0.7740\pm{0.46\%}}$ & $0.7481\pm{0.22\%}$ & ${\bf 0.7768\pm{0.32\%}}$\\
			& $200$ & $0.6697\pm{0.25\%}$ & $0.7039\pm{0.50\%}$ & $0.7061\pm{0.33\%}$ & ${\bf 0.7763\pm{0.42\%}}$ & $0.7484\pm{0.29\%}$ & ${\bf 0.7765\pm{0.22\%}}$\\
			& $250$ & $0.6702\pm{0.23\%}$ & $0.7025\pm{0.29\%}$ & $0.7111\pm{0.37\%}$ & ${\bf 0.7765\pm{0.32\%}}$ & $0.7483\pm{0.21\%}$ & ${\bf 0.7760\pm{0.22\%}}$\\
      \bottomrule
    \end{tabular*}
  \end{threeparttable}
\end{table}

\begin{figure}[htp]	
	\centerline{
		\subfigure[\tiny{AdaGrad}]{\includegraphics[width=.16\textwidth]{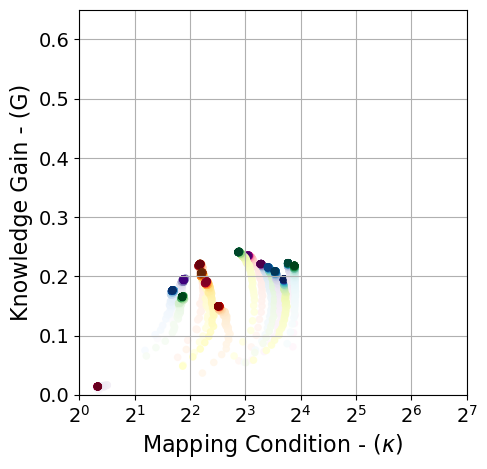}}
		\subfigure[\tiny{RMSProp}]{\includegraphics[width=.16\textwidth]{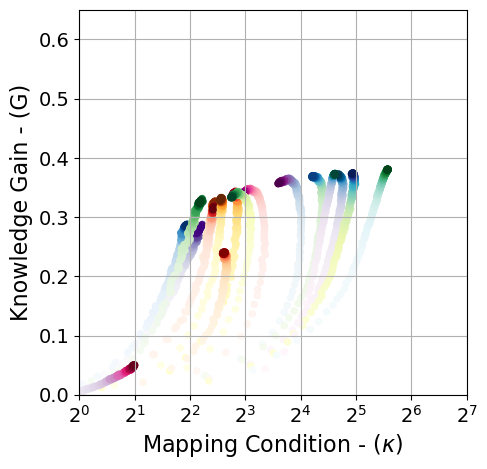}}
		\subfigure[\tiny{AdaM}]{\includegraphics[width=.16\textwidth]{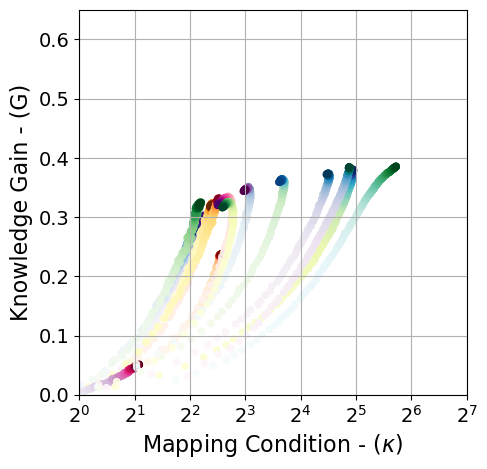}}
		\subfigure[\tiny{AdaBound}]{\includegraphics[width=.16\textwidth]{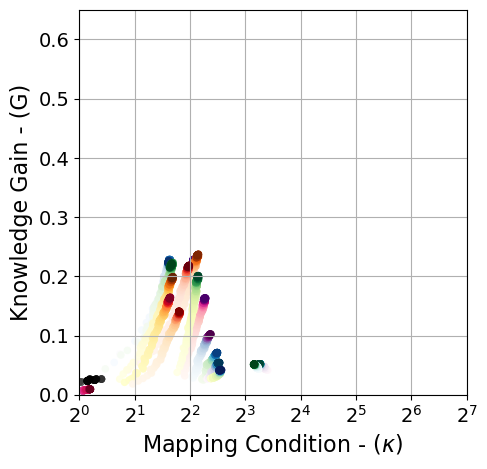}}
		\subfigure[\tiny{SGD-StepLR}]{\includegraphics[width=.16\textwidth]{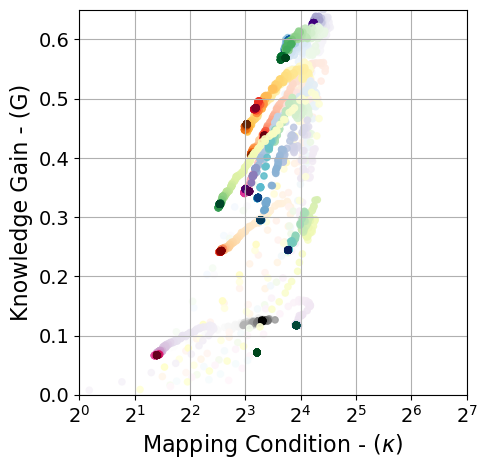}}
		\subfigure[\tiny{SGD-1CycleLR}]{\includegraphics[width=.16\textwidth]{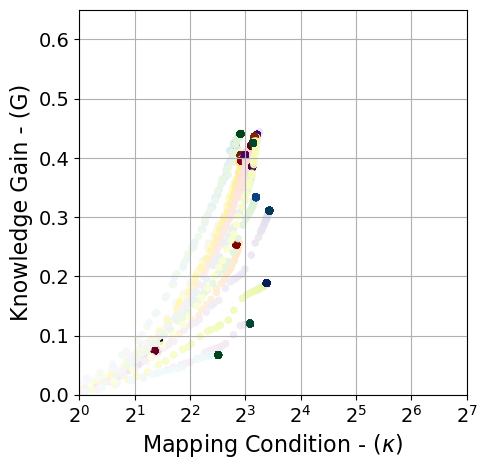}}
	}
	\centerline{
		\subfigure[\tiny{AdaS$^{\beta=0.800}$}]{\includegraphics[width=.16\textwidth]{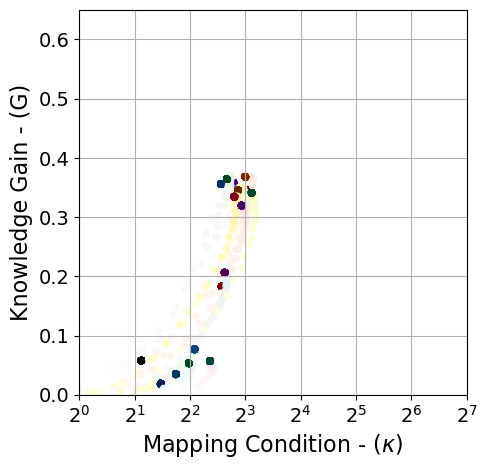}}
		\subfigure[\tiny{AdaS$^{\beta=0.850}$}]{\includegraphics[width=.16\textwidth]{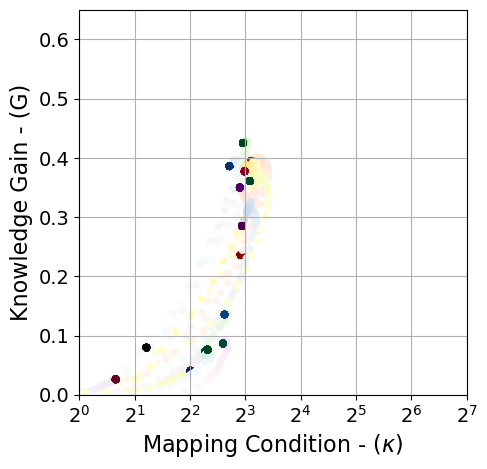}}
		\subfigure[\tiny{AdaS$^{\beta=0.900}$}]{\includegraphics[width=.16\textwidth]{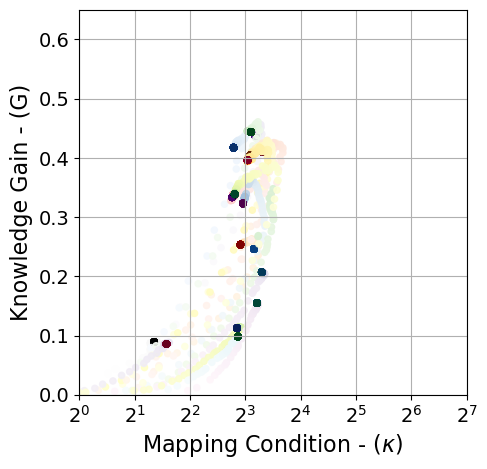}}
		\subfigure[\tiny{AdaS$^{\beta=0.925}$}]{\includegraphics[width=.16\textwidth]{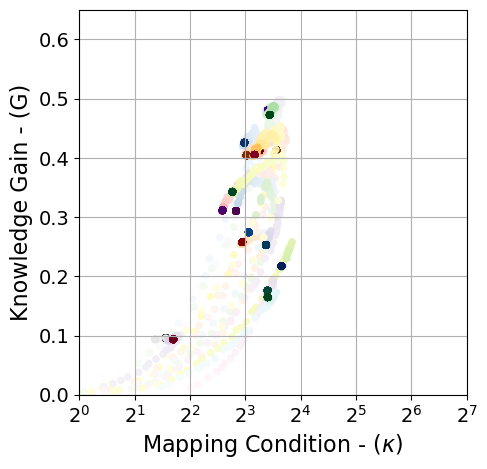}}
		\subfigure[\tiny{AdaS$^{\beta=0.950}$}]{\includegraphics[width=.16\textwidth]{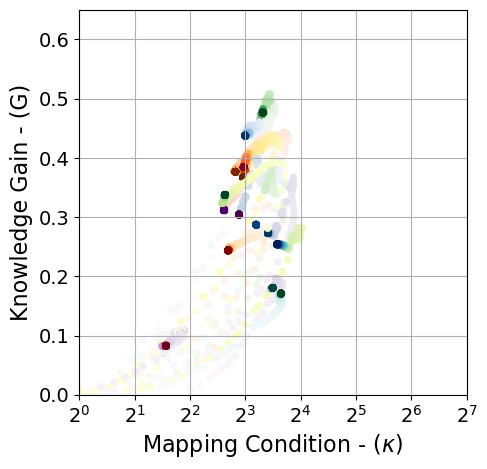}}
		\subfigure[\tiny{AdaS$^{\beta=0.975}$}]{\includegraphics[width=.16\textwidth]{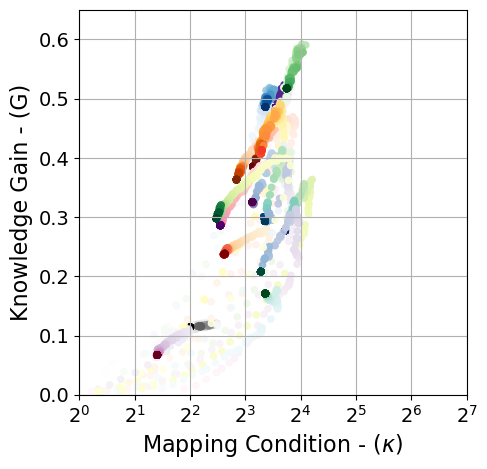}}
	}
	\caption{Evolution of knowledge gain versus mapping condition across different training epochs using ResNet34 on CIFAR10. The transition of color shades correspond to different convolution blocks. The transparency of scatter plots corresponds to the epoch convergence--the higher transparency is inversely related to the training epoch. For complete results of different optimizers and models, please refer to Appendix-B.}
	\label{fig_mapping_condition}
\end{figure}

\subsection{Dependence of the Quality of Network Training to Knowledge Gain}
Both concepts of knowledge gain $G$ and mapping condition $\kappa$ can be used to prob within an intermediate layer of a CNN and quantify the quality of training with respect to different parameter settings. Such quantization do not require test or evaluation data where one can directly measure the  \textit{``expected performance''} of the network throughout the training updates. Our first observation here is made by linking the knowledge gain measure to the relative success of each method in the test accuracy performance. For instance, by raising the gain factor $\beta$ in AdaS, the deeper layers of CNN eventually gain further knowledge as shown in Figure \ref{fig_mapping_condition}. This directly affects the success in test performance results. Also, deploying different optimizers yields different behavior in knowledge gain obtained in different layers of the CNN. Table \ref{tab_acc_loss_G_kappa} lists all four numerical measurements of Test Accuracy, Training Loss, Knowledge Gain, and Mapping Condition for different optimizers. Note the rank order correlation of knowledge gain and test accuracy. Although for both RMSProp and AdaM the knowledge gains are high, however, the Mapping Conditions are also high which deteriorates the overall performance of the network.

\begin{table}[htp]
  \tiny
  \caption{Performance of ResNet34 on CIFAR10 dataset reported at final training epoch for each optimizer. The average scores are reported for $G$ and $\kappa$ across all convolutional blocks.}
  \label{tab_acc_loss_G_kappa}
  \centering  
  \begin{threeparttable}
    \tiny
    \begin{tabular*}{\textwidth}
      {p{2cm}p{0.65cm}p{0.65cm}p{0.65cm}p{0.65cm}p{0.65cm}p{0.65cm}p{0.65cm}p{0.65cm}p{0.65cm}p{0.65cm}}    
      \toprule      
      & \rothead{AdaGrad} & \rothead{RMSProp} & \rothead{AdaM} & \rothead{AdaBound} & \rothead{SGD-1CycleLR} & \rothead{SGD-StepLR} & \rothead{AdaS$^{\beta=0.80}$} & \rothead{AdaS$^{\beta=0.85}$} & \rothead{AdaS$^{\beta=0.90}$} & \rothead{AdaS$^{\beta=0.95}$} \\      
      \toprule
      Test Accuracy  & $0.9150$ & $0.9290$ & $0.9322$ & $0.9274$ & $0.9413$ & $0.9543$ & $0.9302$ & $0.9370$ & $0.9446$ & $0.9516$ \\ \hline
			Training Loss & $1.25\text{e-}3$ & $6.35\text{e-}3$ & $5.66\text{e-}3$ & $2.30\text{e-}3$ & $4.90\text{e-}3$ & $9.70\text{e-}4$ & $6.10\text{e-}3$ & $2.33\text{e-}3$ & $1.27\text{e-}3$ & $8.69\text{e-}4$  \\ \hline
			Knowledge Gain ($G$)  & $0.1790$ & $0.3011$ & $0.2973$ & $0.1288$ & $0.2965$ & $0.3180$ & $0.2063$ & $0.2467$ & $0.2791$ & $0.2938$ \\ \hline
			Mapping Condition ($\kappa$)  & $7.482$ & $18.124$ & $18.484$ & $4.833$ & $7.478$ & $10.086$ & $5.524$ & $6.661$ & $7.839$ & $8.783$ \\
      \bottomrule\\
    \end{tabular*}
  \end{threeparttable}
\end{table}

Our second observation here is made by studying the effect of mapping condition and how it relates to the possible lack of generalizability of each optimizer. Although adaptive optimizers (e.g. RMSProp and AdaM) yield lower training loss, they over-fit perturbing features (mainly caused by incomplete second order statistic measure e.g. diagonal Hessian approximation) and accordingly hamper their generalization \cite{wilson2017marginal}. We suspect such unwanted phenomena is related to the mapping condition within CNN layers. In fact, a mixture of both  average $\kappa$ and average $G$ can help to better realize how well each optimizer can be generalized for training/testing evaluations.

We conclude by identifying that an ideal optimizer would yield $G \rightarrow 1$ and $\kappa \rightarrow 1$ across all layers within a CNN. We highlight that increases in $\kappa$ correlates to greater disentanglement between intermediate input and output layers, hampering the flow of information. Further, we identify that increases in knowledge gain strengthen the carriage of information through the network which enables greater performance.

\section{Conclusion}\label{sec_conclusion}
We have introduced a new adaptive method called AdaS to solve the issue of combined fast convergence and high precision performance of SGD in deep neural networks--all in a unified optimization framework. The method combines the low-rank approximation framework in each convolution layer and identifies how much knowledge is gained in the progression of epoch training and adapts the SGD learning rate proportionally to the rate of change in knowledge gain. AdaS adds minimal computational overhead on the regular SGD algorithm and accordingly provides a well generalized framework to trade off between convergence speed and performance results. Furthermore, AdaS provides an optimization framework which suggests the validation data is no longer required and the stopping criteria for training can be obtained directly from the training loss. Empirical evaluations reveal the possible existence of a lower-bound on SGD step-size that can monotonically increase the knowledge gain independently to each network convolution layer and accordingly improve the overall performance. AdaS is capable of significant improvements in generalization over traditional adaptive methods (i.e. AdaM) while maintaining their rapid convergence characteristics. We highlight that these improvements come through the application of AdaS to simple SGD with momentum. We further identify that since AdaS adaptively tunes the learning rates $\eta(t,\ell)$ independently to all convolutional blocks, it can be deployed with adaptive methods such as AdaM, replacing the traditional scheduling techniques. We postulate that such deployments of AdaS with adaptive gradient updates could introduce greater robustness to initial learning rate choice and leave this exploration as future work. Finally, we emphasize that, without loss of generality, AdaS can be deployed on fully-connected networks, where the weight matrices can be directly fed into the low-rank factorization for metric evaluations.

\section*{Broader Impact}



The content of this research is of broad interest to both researchers and practitioners of computer science and engineering for training deep learning models in machine learning. The method proposed in this paper introduces a new optimization tool that can be adopted for training variety of models such as Convolutional Neural Network (CNN). The proposed optimizer has strong generalizability that include both fast convergence speed and also achieve superior performance compared to the existing off-the-shelf optimizers. The method further introduces a new concept that measures how well the CNN model is trained by probing in different layers of the network and obtain a quality measure for training. This metric can be of broad interest to computer scientists and engineers to develop efficient models that can be tailored on specific applications and dataset.




{\small
\bibliographystyle{ieee_fullname}
\bibliography{egbib}
}
\pagebreak
\appendix
\section*{Appendix-A: Proof of Theorems}

The Proofs \ref{proof_p2} and \ref{proof_p1} correspond  to the proof of Theorem \ref{theorem_knowledge_gain_SGD} for $p=2$ and $p=1$, respectively.
\begin{proof}{(Theorem \ref{theorem_knowledge_gain_SGD})}\label{proof_p2}
For simplicity of notations, we assume the following replacements ${\bf A} = {\bf\Phi}^{k_a}$, ${\bf B} = \sum^{k_b}_{k=k_a}\nabla\tilde{f}_k({\bf\Phi}^{k})$, and ${\bf C} = {\bf\Phi}^{k_b}$. So, the SGD update in (\ref{eq_AdaS_2}) changes to ${\bf C}={\bf A}-\eta{\bf B}$. Using the Definition \ref{definition_knowledge_gain}, the knowledge gain of matrix ${\bf C}$ (assumed to be a column matrix $N\leq M$) is expressed by
\begin{equation}
G({\bf C}) = 
\frac{1}{N\sigma^2_{1}({\bf C})}\sum^{N^{\prime}}_{i=1}{\sigma^2_{i}({\bf C})} =
\frac{1}{N||{\bf C}||^2_2}\trace{{\bf C}^T{\bf C}}.
\label{eq_AdaS_4}
\end{equation}
An upper-bound of first singular value can be calculated by first recalling its equivalence to $\ell_2$-norm and then applying the triangular inequality
\begin{equation}
\sigma^2_{1}({\bf C}) = ||{\bf C}||^2_2 = ||{\bf A}-\eta{\bf B}||^2_2 \leq ||{\bf A}||^2_2 + \eta^2||{\bf B}||^2_2 + 2\eta||{\bf A}||_2||{\bf B}||_2.
\label{eq_AdaS_5}
\end{equation}
By substituting (\ref{eq_AdaS_5}) in (\ref{eq_AdaS_4}) and expanding the terms in trace, a lower bound of ${\bf C}$ is given by
\begin{equation}
G({\bf C}) \geq \frac{1}{N\gamma}\left[\trace{{\bf A}^T{\bf A}}-2\eta\trace{{\bf A}^T{\bf B}}+\eta^2\trace{{\bf B}^T{\bf B}}\right],
\label{eq_AdaS_6}
\end{equation}
where, $\gamma=||{\bf A}||^2_2 + \eta^2||{\bf B}||^2_2 + 2\eta||{\bf A}||_2||{\bf B}||_2$. The latter inequality can be revised to
\begin{align}
\begin{array}{l}
G({\bf C}) \geq \frac{1}{N\gamma}
\left[\left(1-\frac{\gamma}{||{\bf A}||^2_2}+\frac{\gamma}{||{\bf A}||^2_2}\right)\trace{{\bf A}^T{\bf A}}-2\eta\trace{{\bf A}^T{\bf B}}+\eta^2\trace{{\bf B}^T{\bf B}}\right] \\
~~~~~=\frac{1}{N\gamma}
\left[\frac{\gamma}{||{\bf A}||^2_2}\trace{{\bf A}^T{\bf A}}+\left(1-\frac{\gamma}{||{\bf A}||^2_2}\right)\trace{{\bf A}^T{\bf A}}-2\eta\trace{{\bf A}^T{\bf B}}+\eta^2\trace{{\bf B}^T{\bf B}}\right]\\
~~~~~=G({\bf A}) + \frac{1}{N\gamma}
\left[\underbrace{\left(1-\frac{\gamma}{||{\bf A}||^2_2}\right)\trace{{\bf A}^T{\bf A}}-2\eta\trace{{\bf A}^T{\bf B}}+\eta^2\trace{{\bf B}^T{\bf B}}}_{D}\right].
\end{array}
\label{eq_AdaS_7}
\end{align}
Therefore, the bound in (\ref{eq_AdaS_7}) is revised to
\begin{equation}
G({\bf C}) - G({\bf A}) \geq \frac{1}{N\gamma}D.
\label{eq_AdaS_8}
\end{equation}
The monotonicity of the knowledge gain in (\ref{eq_AdaS_8}) is guaranteed if $D\geq{0}$. The remaining term $D$ can be expressed as a quadratic function of $\eta$
\begin{equation}
D(\eta) = \left[\trace{{\bf B}^T{\bf B}}-\frac{||{\bf B}||^2_2}{||{\bf A}||^2_2}\trace{{\bf A}^T{\bf A}}\right]\eta^2 - \left[2\trace{{\bf A}^T{\bf B}} + 2\frac{||{\bf B}||_2}{||{\bf A}||_2}\trace{{\bf A}^T{\bf A}}\right]\eta
\label{eq_AdaS_9}
\end{equation}
where, the condition for $D(\eta)\geq{0}$ in (\ref{eq_AdaS_9}) is 
\begin{equation}
\eta\geq \max\left\{2\frac{\trace{{\bf A}^T{\bf B}} + \frac{||{\bf B}||_2}{||{\bf A}||_2}\trace{{\bf A}^T{\bf A}}}{\trace{{\bf B}^T{\bf B}}-\frac{||{\bf B}||^2_2}{||{\bf A}||^2_2}\trace{{\bf A}^T{\bf A}}},~0\right\}.
\label{eq_AdaS_10}
\end{equation}
Hence, given the lower bound in (\ref{eq_AdaS_10}) it will guarantee the monotonicity of the knowledge gain through the update scheme ${\bf C}={\bf A}-\eta{\bf B}$.

Our final inspection is to check if the substitution of step-size (\ref{eq_AdaS_3}) in (\ref{eq_AdaS_8}) would still hold the inequality condition in (\ref{eq_AdaS_8}). Followed by the substitution, the inequality should satisfy
\begin{equation}
\eta \geq \zeta\frac{1}{N\gamma}D.
\label{eq_AdaS_11}
\end{equation}
We have found that $D(\eta)\geq{0}$ for some lower bound in (\ref{eq_AdaS_10}), where the inequality in (\ref{eq_AdaS_11}) also holds from some $\zeta\geq{0}$ and the proof is done.
\end{proof}

\begin{proof}{(Theorem \ref{theorem_knowledge_gain_SGD})}\label{proof_p1}
Following similar notation in Proof \ref{proof_p2}, the knowledge gain of matrix ${\bf C}$ is expressed by
\begin{equation}
G({\bf C}) = 
\frac{1}{N\sigma_{1}({\bf C})}\sum^{N^{\prime}}_{i=1}{\sigma_{i}({\bf C})}.
\label{eq_AdaS_12}
\end{equation}
By stacking all singular values in a vector form (and recall from $\ell_1$ and $\ell_2$ norms inequality)
\begin{equation}
\left[\sum^{N^{\prime}}_{i=1}{\sigma_{i}({\bf C})}\right]^2 =
||\underline{\sigma}({\bf C})||^2_1\geq||\underline{\sigma}({\bf C})||^2_2 =
\sum^{N^{\prime}}_{i=1}{\sigma^2_{i}({\bf C})} = 
\trace{{\bf C}^T{\bf C}},
\notag
\end{equation}
and by substituting the matrix composition ${\bf C}$, the following inequality holds
\begin{equation}
\left[\sum^{N^{\prime}}_{i=1}{\sigma_{i}({\bf C})}\right]^2 \geq
\trace{{\bf A}^T{\bf A}} - 2\eta\trace{{\bf A}^T{\bf B}}+\eta^2\trace{{\bf B}^T{\bf B}}.
\label{eq_AdaS_13}
\end{equation}

An upper-bound of first singular value can be calculated by recalling its equivalence to $\ell_2$-norm and triangular inequality as follows
\begin{equation}
\sigma^2_{1}({\bf C}) = ||{\bf C}||^2_2 = ||{\bf A}-\eta{\bf B}||^2_2 \leq ||{\bf A}||^2_2 +  2\eta||{\bf A}||_2||{\bf B}||_2 + \eta^2||{\bf B}||^2_2.
\label{eq_AdaS_14}
\end{equation}
By substituting the lower-bound (\ref{eq_AdaS_13}) and upper-bound (\ref{eq_AdaS_14}) into (\ref{eq_AdaS_12}), a lower bound of knowledge gain is given by
\begin{equation}
G^2({\bf C}) \geq \frac{1}{N^2\gamma}\left[\trace{{\bf A}^T{\bf A}} - 2\eta\trace{{\bf A}^T{\bf B}}+\eta^2\trace{{\bf B}^T{\bf B}}\right],
\notag
\end{equation}
where $\gamma = ||{\bf A}||^2_2 +  2\eta||{\bf A}||_2||{\bf B}||_2 + \eta^2||{\bf B}||^2_2$. The latter inequality can be revised to
\begin{equation}
G^2({\bf C}) \geq \frac{1}{N^2\gamma}[\frac{N^{\prime\prime}\gamma}{||{\bf A}||^2_2}\trace{{\bf A}^T{\bf A}} +
\underbrace{(1-\frac{N^{\prime\prime}\gamma}{||{\bf A}||^2_2})\trace{{\bf A}^T{\bf A}} - 2\eta\trace{{\bf A}^T{\bf B}}+\eta^2\trace{{\bf B}^T{\bf B}}}_{D}],
\label{eq_AdaS_15}
\end{equation}
where, the lower bound of the first summand term is given by
\begin{equation}
\frac{N^{\prime\prime}\gamma}{||{\bf A}||^2_2}\trace{{\bf A}^T{\bf A}} =
\frac{N^{\prime\prime}\gamma}{||{\bf A}||^2_2}\sum^{N^{\prime\prime}}_{i=1}{\sigma^2_{i}({\bf A})} = 
\frac{N^{\prime\prime}\gamma}{\sigma^2_{1}({\bf A})}||\underline{\sigma}({\bf A})||^2_2 \geq
\frac{\gamma}{\sigma^2_{1}({\bf A})}||\underline{\sigma}({\bf A})||^2_1 =
\gamma{N^2}G^2({\bf A}).
\notag
\end{equation}
Therefore, the bound in (\ref{eq_AdaS_15}) is revised to
\begin{equation}
G^2({\bf C}) \geq G^2({\bf A}) + \frac{1}{N^2\gamma}D.
\label{eq_AdaS_16}
\end{equation}
Note that $\gamma\geq{0}$ (step-size $\eta\geq{0}$ is always positive) and the only condition for the bound in (\ref{eq_AdaS_16}) to hold is to $D\geq{0}$. Here the remaining term $D$ can be expressed as quadratic function of step-size i.e. $D(\eta) = a\eta^2+b\eta+c$ where
\begin{equation}
a = \trace{{\bf B}^T{\bf B}} - N^{\prime\prime}\frac{||{\bf B}||^2_2}{||{\bf A}||^2_2}\trace{{\bf A}^T{\bf A}},~~~
b = - 2\trace{{\bf A}^T{\bf B}} - N^{\prime\prime}\frac{||{\bf B}||_2}{||{\bf A}||_2},~~~
c = -(N^{\prime\prime}-1)\trace{{\bf A}^T{\bf A}}.
\nonumber
\end{equation}
The quadratic function can be factorized $D(\eta)=(\eta-\eta_1)(\eta-\eta_2)$ where the roots $\eta_1=(-b+\sqrt{\Delta})/2a$ and $\eta_2=(-b-\sqrt{\Delta})/2a$, and $\Delta=b^2-4ac$. Here $c\leq{0}$ and assuming $a\geq{0}$ then $\Delta\geq{0}$. Accordingly, $\eta_1\geq{0}$ and $\eta_2\leq{0}$. For the function $D(\eta)$ to yield a positive value, both factorized elements should be either positive (i.e. $\eta-\eta_1\geq{0}$ and $\eta-\eta_2\geq{0}$) or negative (i.e. $\eta-\eta_1\leq{0}$ and $\eta-\eta_2\leq{0}$). Here, only the positive conditions hold which yield $\eta\geq\eta_1$. The assumption $a\geq{0}$ is equivalent to $\sum^{N^{\prime\prime\prime}}_{i=1}{\sigma^2_i({\bf B})/\sigma^2_1({\bf B})}\geq N^{\prime\prime}\sum^{N^{\prime\prime}}_{i=1}{\sigma^2_i({\bf A})/\sigma^2_1({\bf A})}$. The condition strongly holds for the beginning epochs due to random initialization of weights where the low-rank matrix ${\bf A}$ is indeed an empty matrix at epoch$=0$. By progression epoch training this condition loosens and might not hold. Therefore, the monotonicity of knowledge gain for $p=1$ could be violated in the interim process.
\end{proof}

\section*{Appendix-B: AdaS Ablation Study}
\begin{figure}[htp]
	\centerline{
		\subfigure[\tiny{CIFAR10/VGG16}]{\includegraphics[width=.24\textwidth]{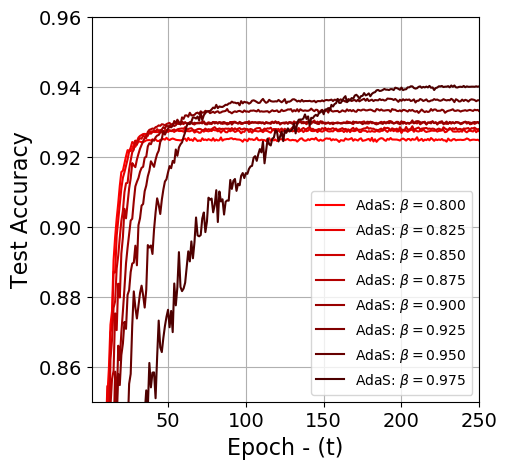}\label{AdaS_test_accuracy_ablative_Study_CIFAR10_VGG16}}
		\subfigure[\tiny{CIFAR10/ResNet34}]{\includegraphics[width=.24\textwidth]{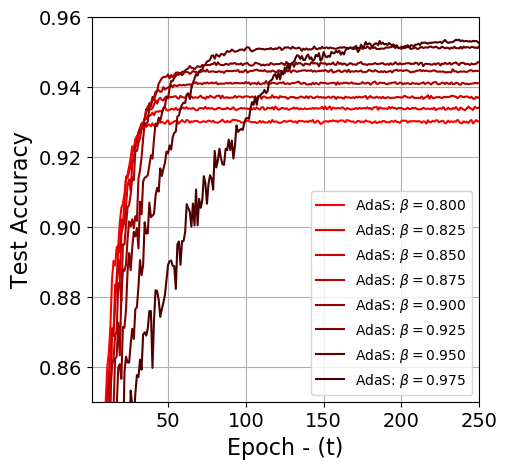}\label{AdaS_test_accuracy_ablative_Study_CIFAR10_ResNet34}}
		\subfigure[\tiny{CIFAR100/VGG16}]{\includegraphics[width=.24\textwidth]{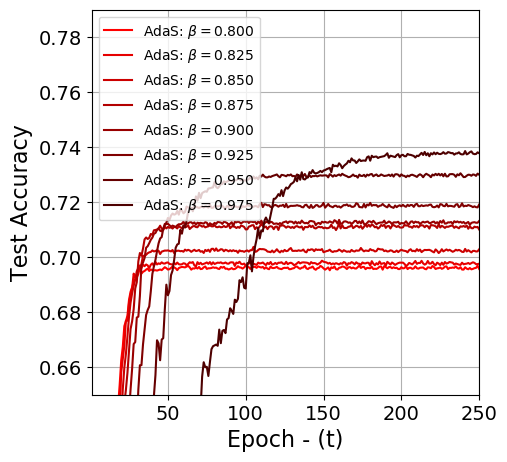}\label{AdaS_test_accuracy_ablative_Study_CIFAR100_VGG16}}
		\subfigure[\tiny{CIFAR100/ResNet34}]{\includegraphics[width=.24\textwidth]{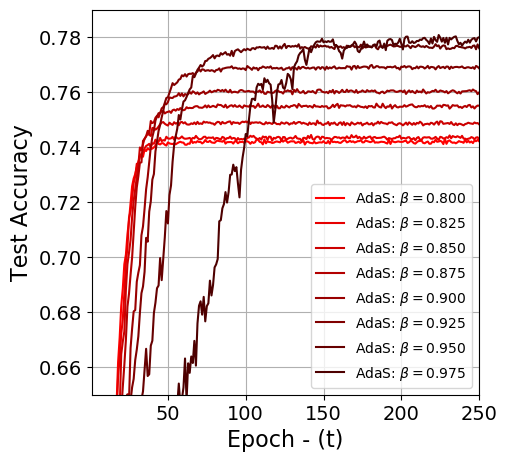}\label{AdaS_test_accuracy_ablative_Study_CIFAR100_ResNet34}}
	}
	\centerline{
		\subfigure[\tiny{CIFAR10/VGG16}]{\includegraphics[width=.24\textwidth]{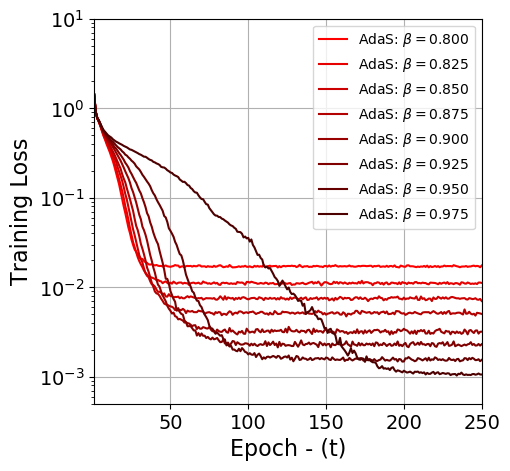}\label{AdaS_training_loss_ablative_Study_CIFAR10_VGG16}}
		\subfigure[\tiny{CIFAR10/ResNet34}]{\includegraphics[width=.24\textwidth]{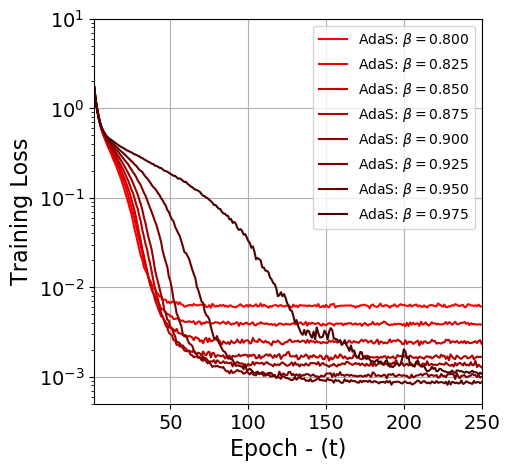}\label{AdaS_training_loss_ablative_Study_CIFAR10_ResNet34}}
		\subfigure[\tiny{CIFAR100/VGG16}]{\includegraphics[width=.24\textwidth]{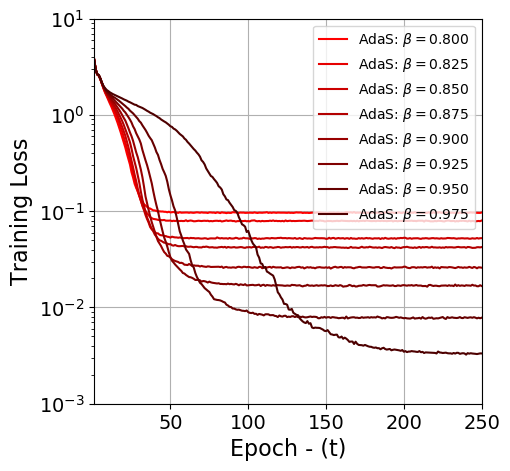}\label{AdaS_training_loss_ablative_Study_CIFAR100_VGG16}}
		\subfigure[\tiny{CIFAR100/ResNet34}]{\includegraphics[width=.24\textwidth]{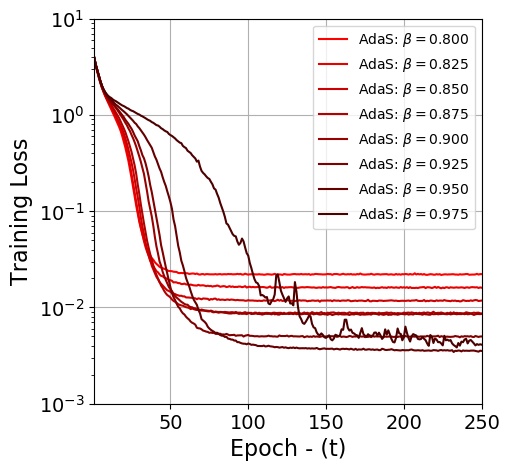}\label{AdaS_training_loss_ablative_Study_CIFAR100_ResNet34}}
	}
	\caption{Ablative study of AdaS momentum rate over two different datasets (i.e. CIFAR10 and CIFAR100) and two CNNs (i.e. VGG16 and ResNet34). Top row corresponds to test-accuracies and bottom row to training-losses.}
	\label{fig_ablative_study_dataset_vs_network}
\end{figure}

\begin{figure}[htp]
	\centerline{
		\subfigure[\tiny{CIFAR10/VGG16/$\beta=0.80$}]{\includegraphics[width=.245\textwidth]{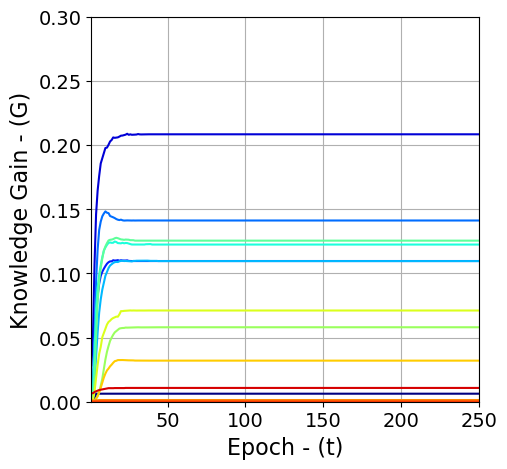}}
		\subfigure[\tiny{CIFAR10/VGG16/$\beta=0.85$}]{\includegraphics[width=.245\textwidth]{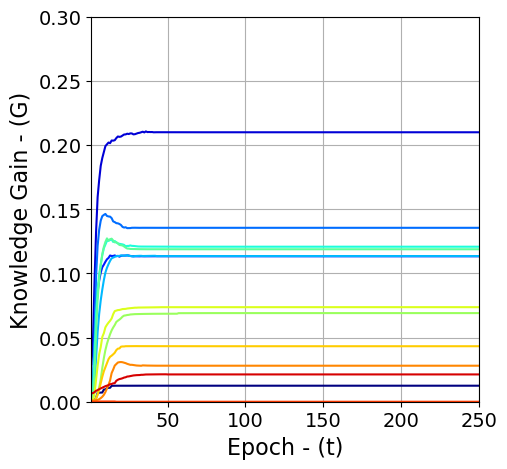}}
		\subfigure[\tiny{CIFAR10/VGG16/$\beta=0.90$}]{\includegraphics[width=.245\textwidth]{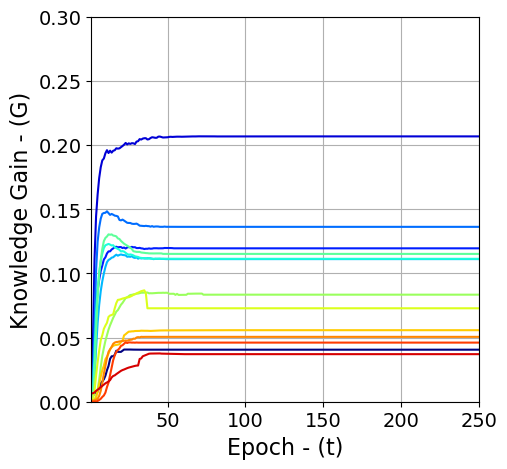}}
		\subfigure[\tiny{CIFAR10/VGG16/$\beta=0.95$}]{\includegraphics[width=.245\textwidth]{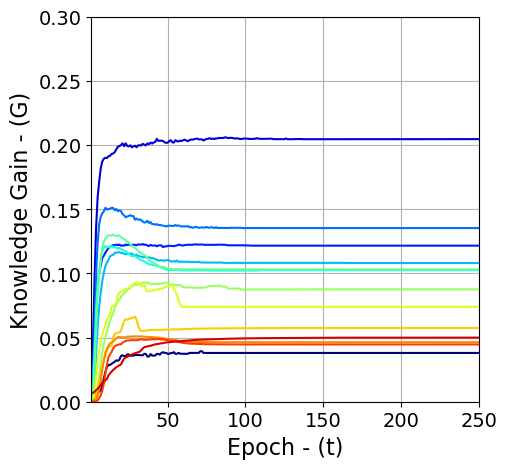}}
	}
	\centerline{
		\subfigure[\tiny{CIFAR10/ResNet34/$\beta=0.80$}]{\includegraphics[width=.245\textwidth]{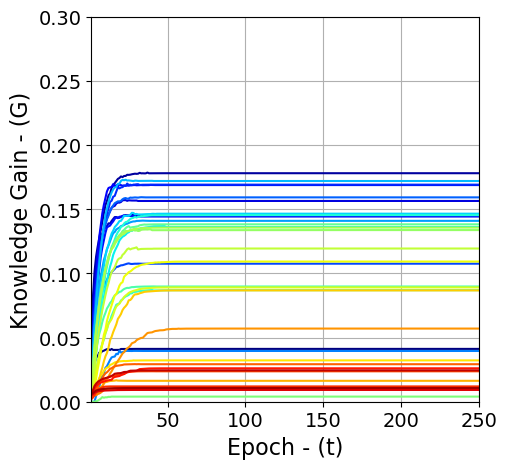}}
		\subfigure[\tiny{CIFAR10/ResNet34/$\beta=0.85$}]{\includegraphics[width=.245\textwidth]{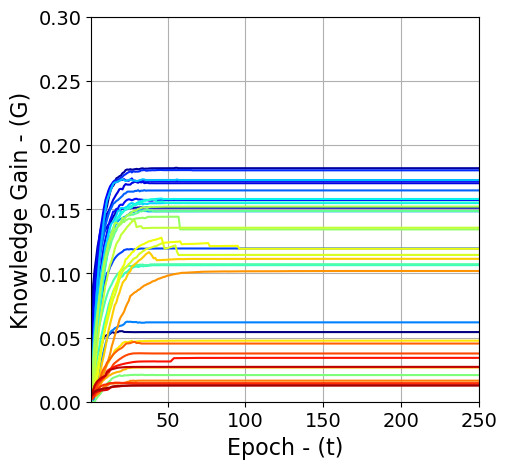}}
		\subfigure[\tiny{CIFAR10/ResNet34/$\beta=0.90$}]{\includegraphics[width=.245\textwidth]{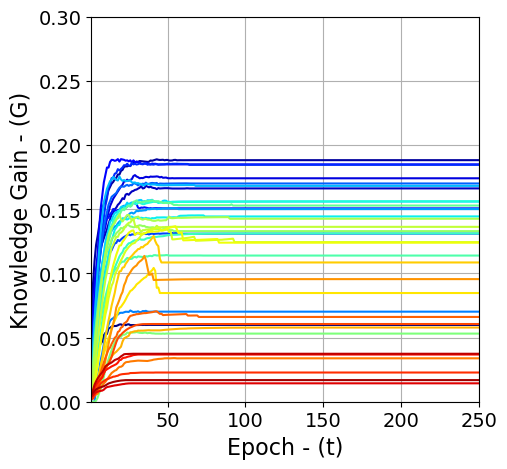}}
		\subfigure[\tiny{CIFAR10/ResNet34/$\beta=0.95$}]{\includegraphics[width=.245\textwidth]{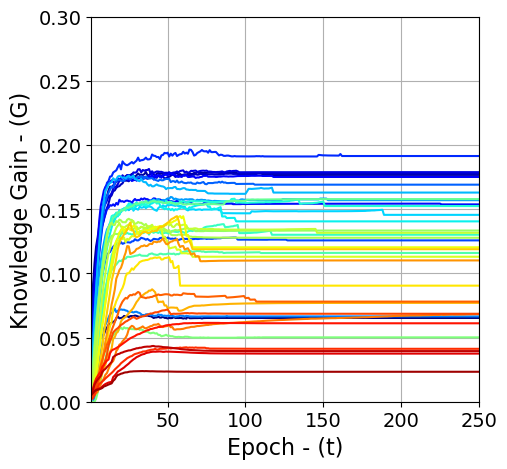}}
	}
	\centerline{
		\subfigure[\tiny{CIFAR100/VGG16/$\beta=0.80$}]{\includegraphics[width=.245\textwidth]{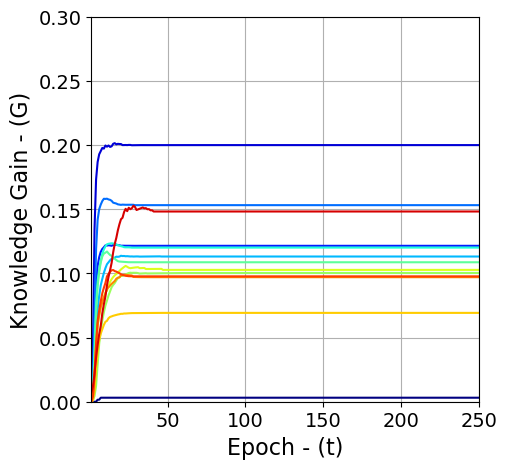}}
		\subfigure[\tiny{CIFAR100/VGG16/$\beta=0.85$}]{\includegraphics[width=.245\textwidth]{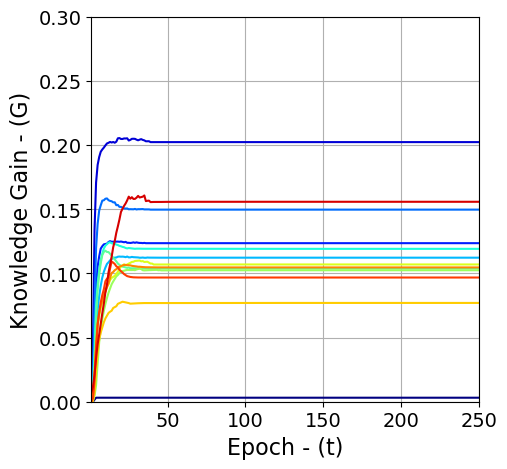}}
		\subfigure[\tiny{CIFAR100/VGG16/$\beta=0.90$}]{\includegraphics[width=.245\textwidth]{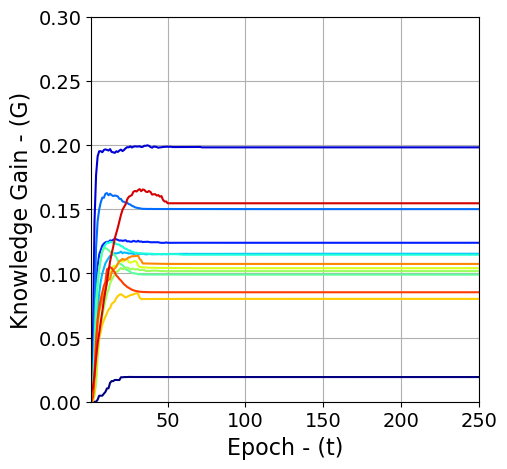}}
		\subfigure[\tiny{CIFAR100/VGG16/$\beta=0.95$}]{\includegraphics[width=.245\textwidth]{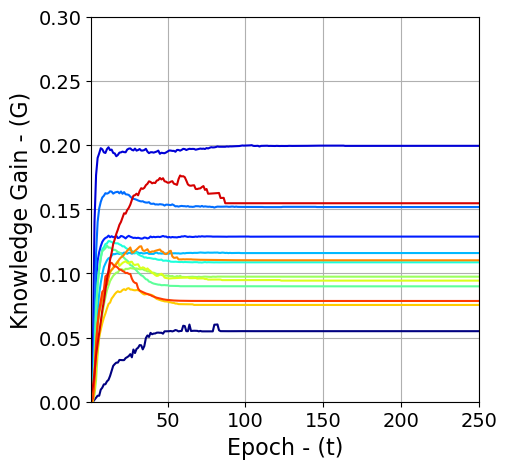}}
	}
	\centerline{
		\subfigure[\tiny{CIFAR100/ResNet34/$\beta=0.80$}]{\includegraphics[width=.245\textwidth]{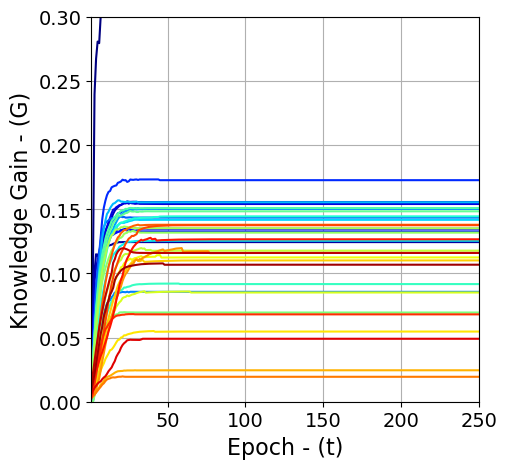}}
		\subfigure[\tiny{CIFAR100/ResNet34/$\beta=0.85$}]{\includegraphics[width=.245\textwidth]{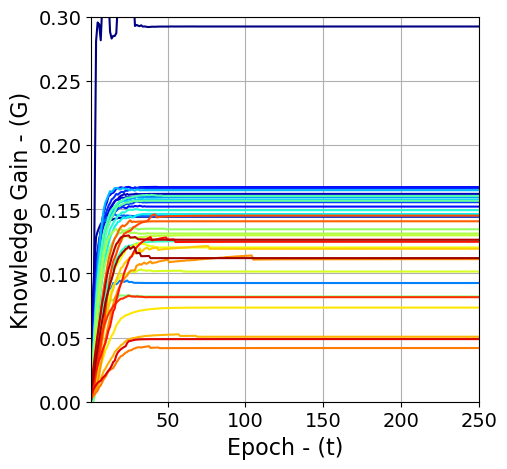}}
		\subfigure[\tiny{CIFAR100/ResNet34/$\beta=0.90$}]{\includegraphics[width=.245\textwidth]{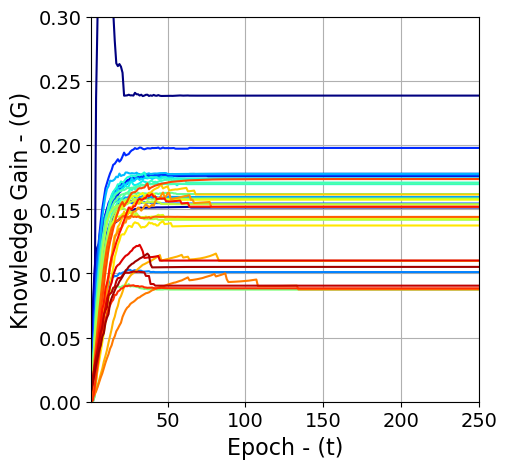}}
		\subfigure[\tiny{CIFAR100/ResNet34/$\beta=0.95$}]{\includegraphics[width=.245\textwidth]{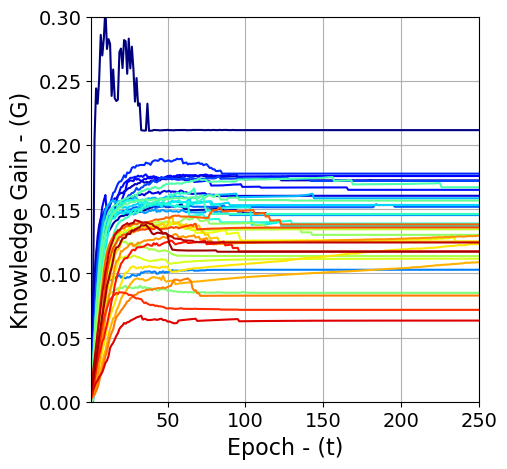}}
	}
	\caption{Ablative study of AdaS momentum rate $\beta$ versus knowledge gain $G$ over two different datasets (i.e. CIFAR10 and CIFAR100) and two CNNs (i.e. VGG16 and ResNet34). The transition in color shades from light to dark lines correspond to first to the end of convolution layers in each network.}
	\label{fig_ablative_study_AdaS_momentum_rate_vs_knowledge_gain}
\end{figure}

\begin{figure}[htp]
	\centerline{
		\subfigure[\tiny{CIFAR10/VGG16/$\beta=0.80$}]{\includegraphics[width=.245\textwidth]{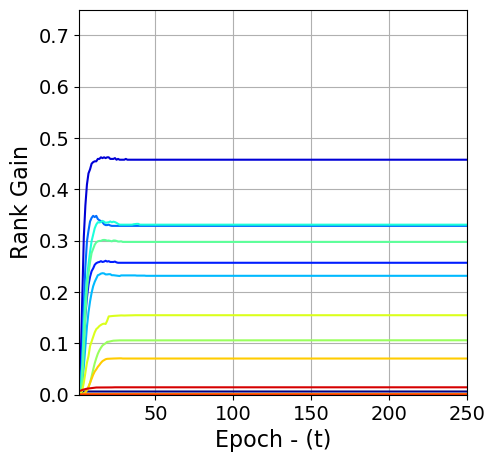}}
		\subfigure[\tiny{CIFAR10/VGG16/$\beta=0.85$}]{\includegraphics[width=.245\textwidth]{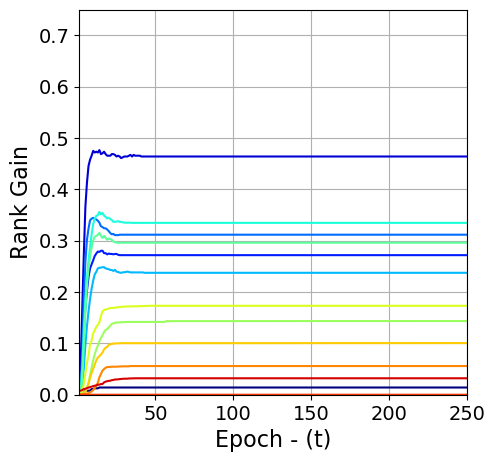}}
		\subfigure[\tiny{CIFAR10/VGG16/$\beta=0.90$}]{\includegraphics[width=.245\textwidth]{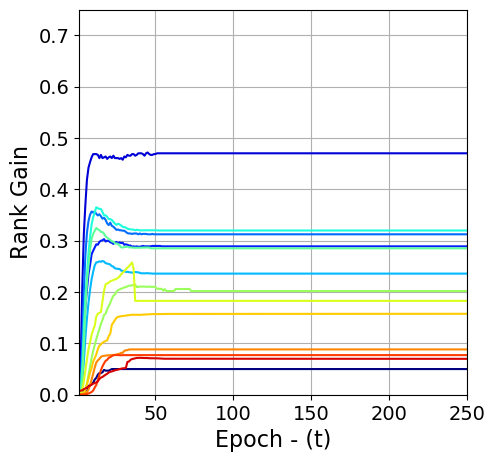}}
		\subfigure[\tiny{CIFAR10/VGG16/$\beta=0.95$}]{\includegraphics[width=.245\textwidth]{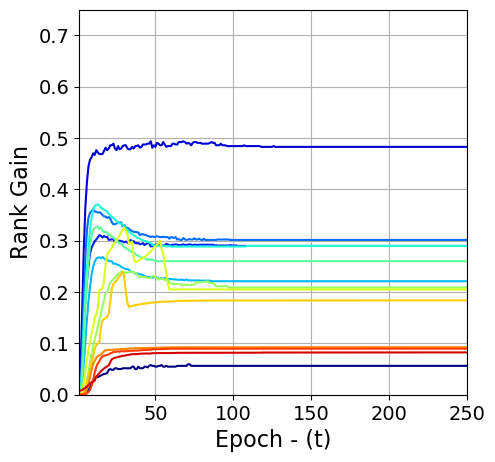}}
	}
	\centerline{
		\subfigure[\tiny{CIFAR10/ResNet34/$\beta=0.80$}]{\includegraphics[width=.245\textwidth]{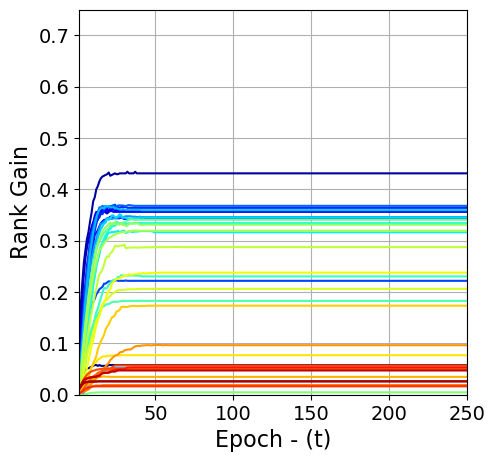}}
		\subfigure[\tiny{CIFAR10/ResNet34/$\beta=0.85$}]{\includegraphics[width=.245\textwidth]{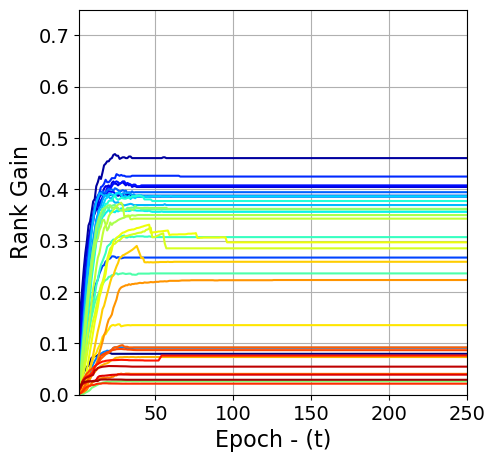}}
		\subfigure[\tiny{CIFAR10/ResNet34/$\beta=0.90$}]{\includegraphics[width=.245\textwidth]{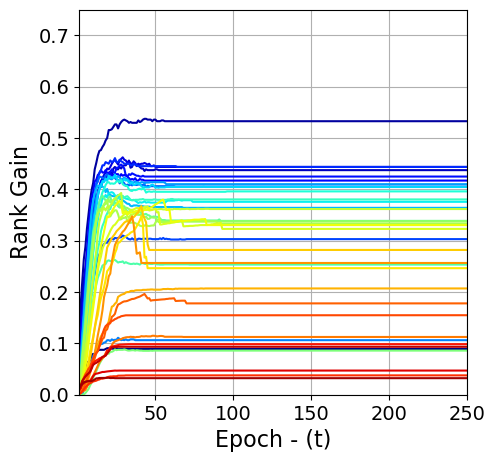}}
		\subfigure[\tiny{CIFAR10/ResNet34/$\beta=0.95$}]{\includegraphics[width=.245\textwidth]{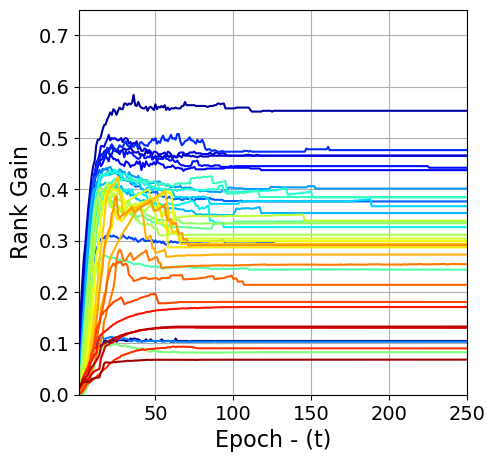}}
	}
	\centerline{
		\subfigure[\tiny{CIFAR100/VGG16/$\beta=0.80$}]{\includegraphics[width=.245\textwidth]{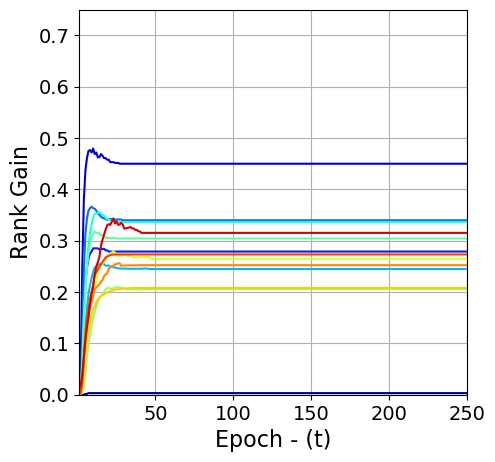}}
		\subfigure[\tiny{CIFAR100/VGG16/$\beta=0.85$}]{\includegraphics[width=.245\textwidth]{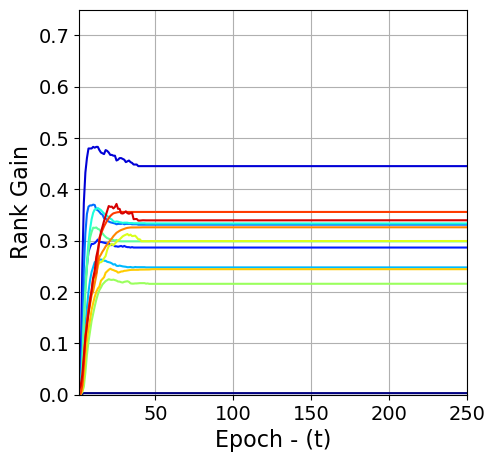}}
		\subfigure[\tiny{CIFAR100/VGG16/$\beta=0.90$}]{\includegraphics[width=.245\textwidth]{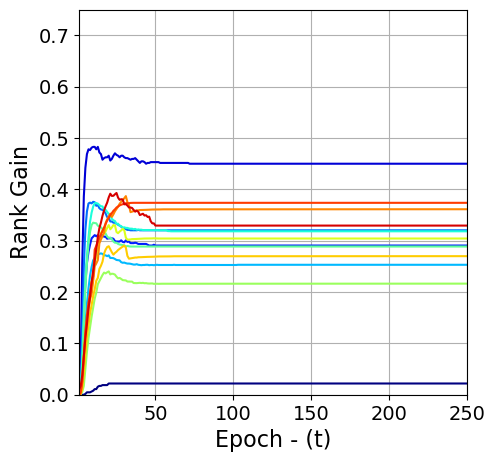}}
		\subfigure[\tiny{CIFAR100/VGG16/$\beta=0.95$}]{\includegraphics[width=.245\textwidth]{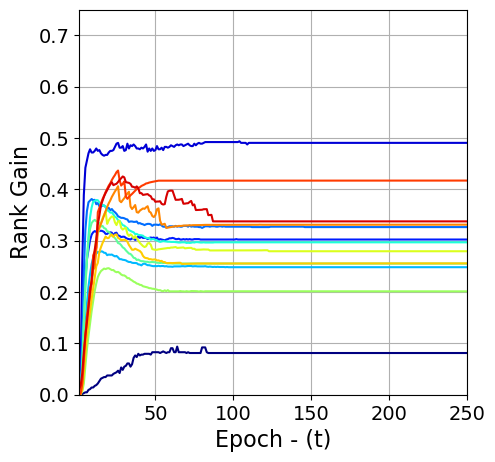}}
	}
	\centerline{
		\subfigure[\tiny{CIFAR100/ResNet34/$\beta=0.80$}]{\includegraphics[width=.245\textwidth]{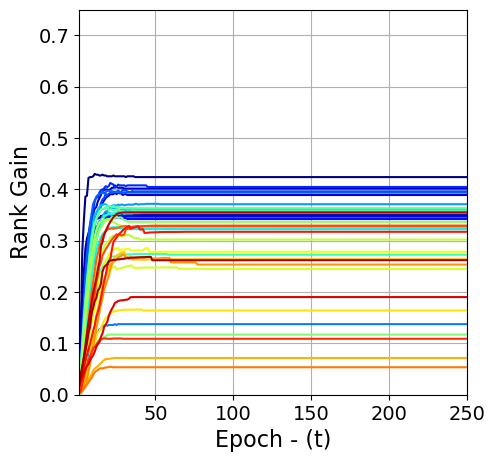}}
		\subfigure[\tiny{CIFAR100/ResNet34/$\beta=0.85$}]{\includegraphics[width=.245\textwidth]{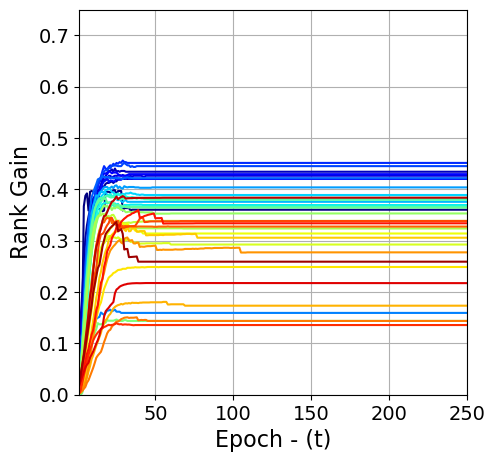}}
		\subfigure[\tiny{CIFAR100/ResNet34/$\beta=0.90$}]{\includegraphics[width=.245\textwidth]{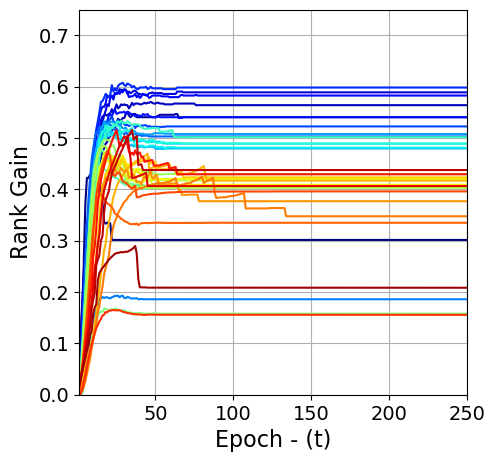}}
		\subfigure[\tiny{CIFAR100/ResNet34/$\beta=0.95$}]{\includegraphics[width=.245\textwidth]{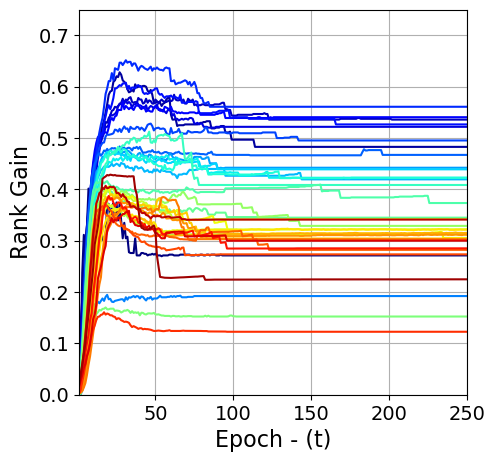}}
	}
	\caption{Ablative study of AdaS momentum rate $\beta$ versus rank gain $\rank{{\bf\hat{\Phi}}}$ over two different datasets (i.e. CIFAR10 and CIFAR100) and two CNNs (i.e. VGG16 and ResNet34). The transition in color shades from light to dark lines correspond to first to the end of convolution layers in each network.}
	\label{fig_ablative_study_AdaS_momentum_rate_vs_rank_gain}
\end{figure}

\begin{figure}[htp]
	\centerline{
		\subfigure[\tiny{CIFAR10/VGG16/$\beta=0.80$}]{\includegraphics[width=.245\textwidth]{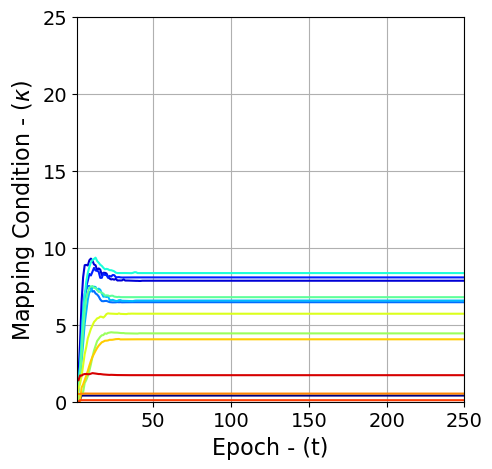}}
		\subfigure[\tiny{CIFAR10/VGG16/$\beta=0.85$}]{\includegraphics[width=.245\textwidth]{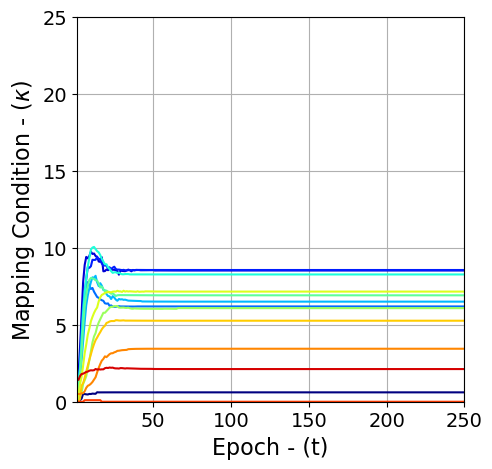}}
		\subfigure[\tiny{CIFAR10/VGG16/$\beta=0.90$}]{\includegraphics[width=.245\textwidth]{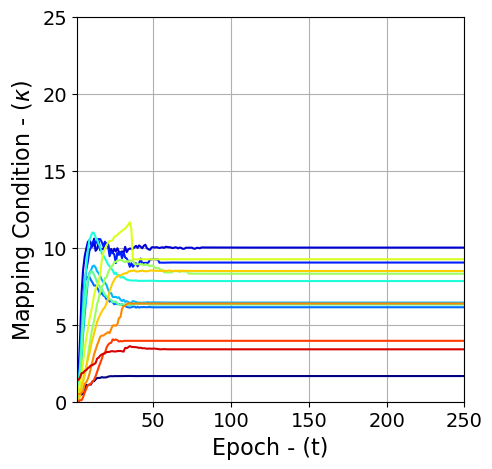}}
		\subfigure[\tiny{CIFAR10/VGG16/$\beta=0.95$}]{\includegraphics[width=.245\textwidth]{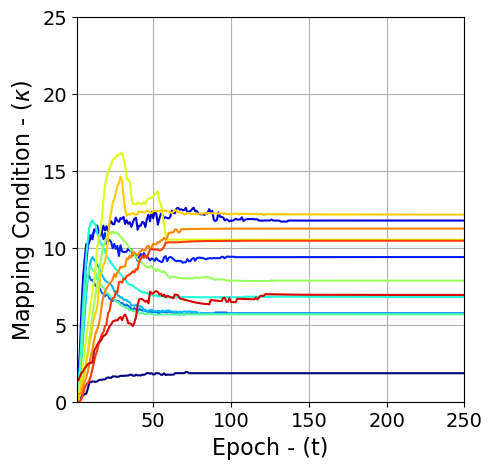}}
	}
	\centerline{
		\subfigure[\tiny{CIFAR10/ResNet34/$\beta=0.80$}]{\includegraphics[width=.245\textwidth]{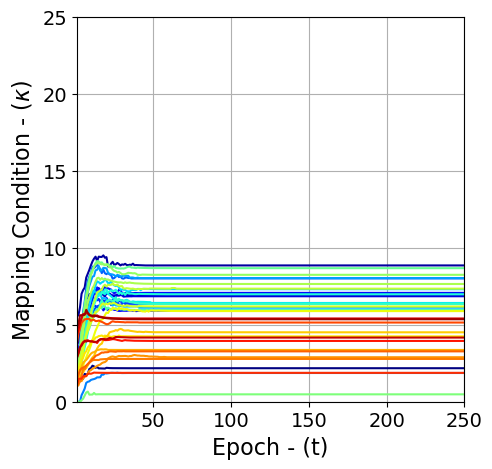}}
		\subfigure[\tiny{CIFAR10/ResNet34/$\beta=0.85$}]{\includegraphics[width=.245\textwidth]{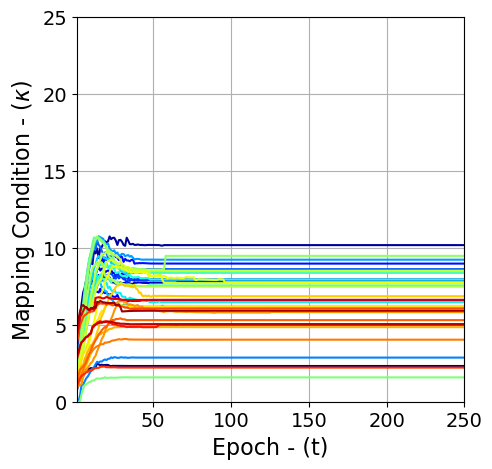}}
		\subfigure[\tiny{CIFAR10/ResNet34/$\beta=0.90$}]{\includegraphics[width=.245\textwidth]{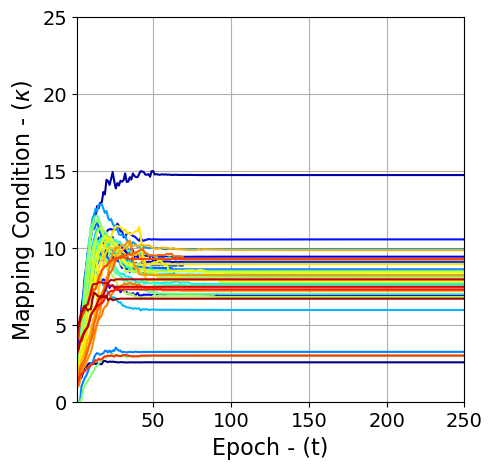}}
		\subfigure[\tiny{CIFAR10/ResNet34/$\beta=0.95$}]{\includegraphics[width=.245\textwidth]{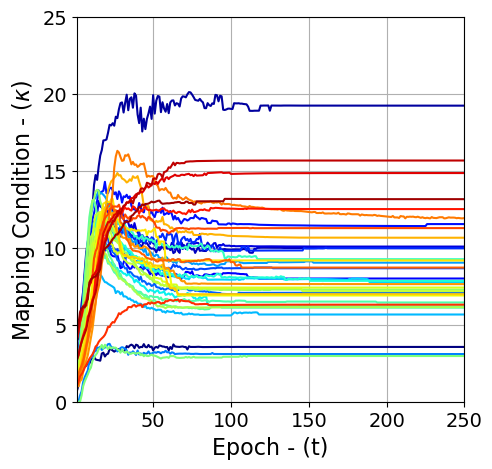}}
	}
	\centerline{
		\subfigure[\tiny{CIFAR100/VGG16/$\beta=0.80$}]{\includegraphics[width=.245\textwidth]{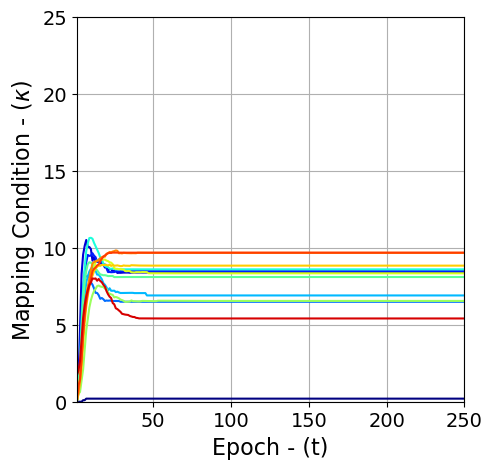}}
		\subfigure[\tiny{CIFAR100/VGG16/$\beta=0.85$}]{\includegraphics[width=.245\textwidth]{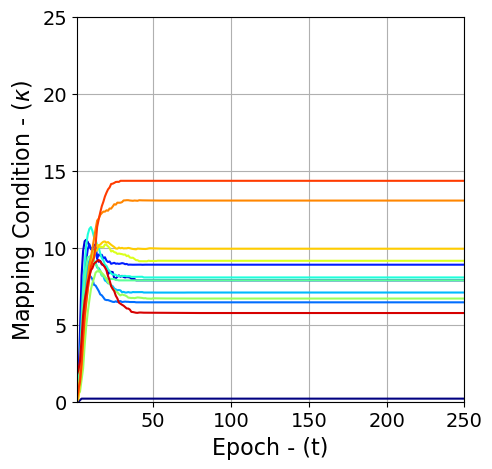}}
		\subfigure[\tiny{CIFAR100/VGG16/$\beta=0.90$}]{\includegraphics[width=.245\textwidth]{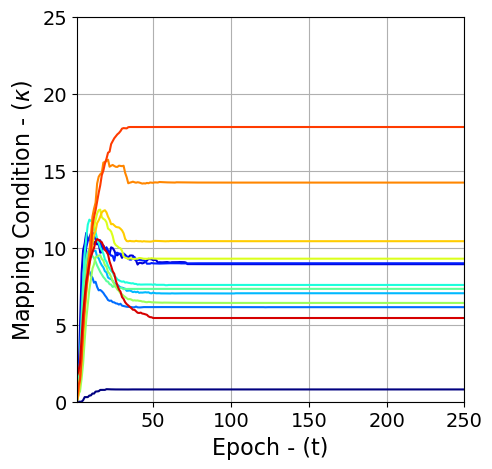}}
		\subfigure[\tiny{CIFAR100/VGG16/$\beta=0.95$}]{\includegraphics[width=.245\textwidth]{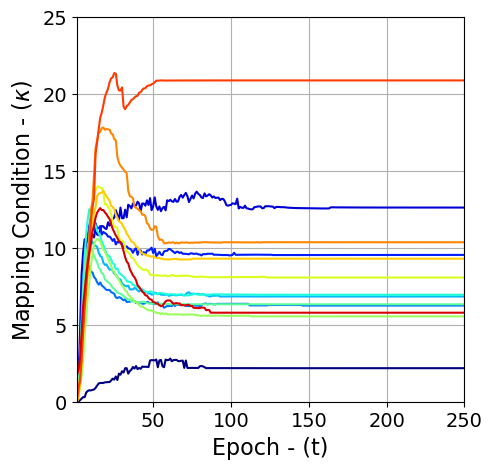}}
	}
	\centerline{
		\subfigure[\tiny{CIFAR100/ResNet34/$\beta=0.80$}]{\includegraphics[width=.245\textwidth]{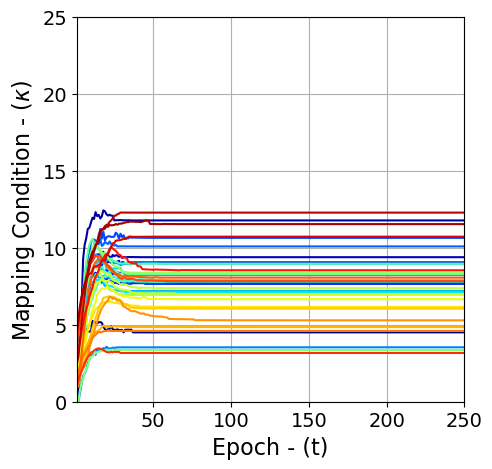}}
		\subfigure[\tiny{CIFAR100/ResNet34/$\beta=0.85$}]{\includegraphics[width=.245\textwidth]{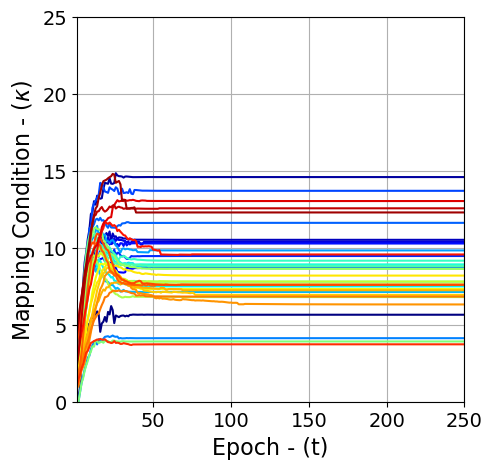}}
		\subfigure[\tiny{CIFAR100/ResNet34/$\beta=0.90$}]{\includegraphics[width=.245\textwidth]{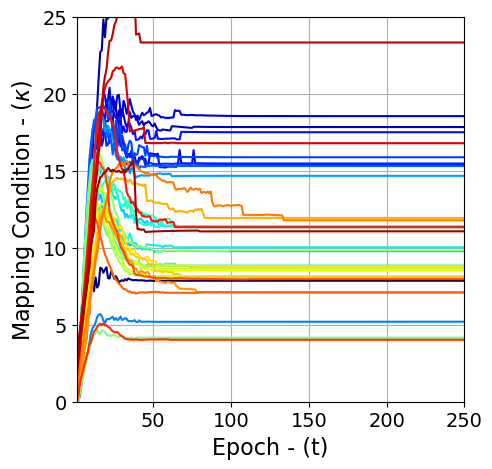}}
		\subfigure[\tiny{CIFAR100/ResNet34/$\beta=0.95$}]{\includegraphics[width=.245\textwidth]{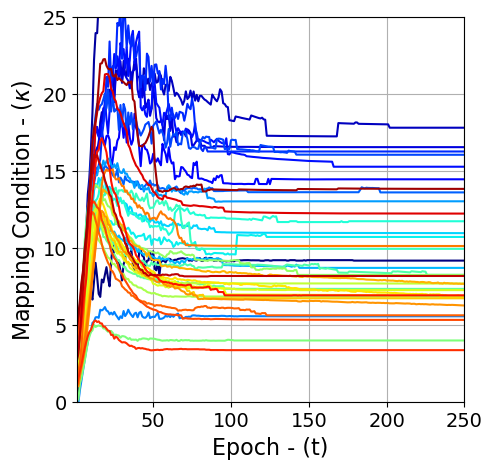}}
	}
	\caption{Ablative study of AdaS momentum rate $\beta$ versus mapping condition $\kappa$ over two different datasets (i.e. CIFAR10 and CIFAR100) and two CNNs (i.e. VGG16 and ResNet34). The transition in color shades from light to dark lines correspond to first to the end of convolution layers in each network.}
	\label{fig_ablative_study_AdaS_momentum_rate_vs_mapping_condition}
\end{figure}

\begin{figure}[htp]
	\centerline{
		\subfigure[\tiny{CIFAR10/VGG16/$\beta=0.80$}]{\includegraphics[width=.245\textwidth]{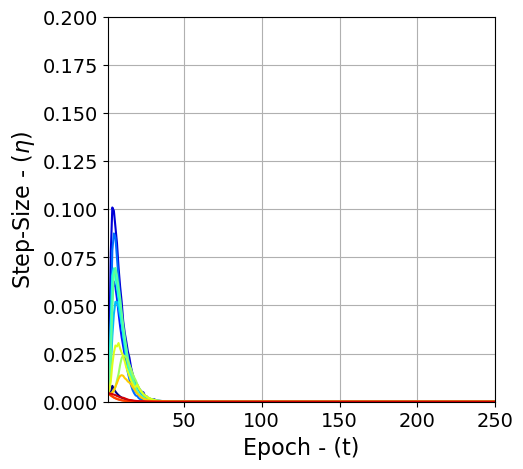}}
		\subfigure[\tiny{CIFAR10/VGG16/$\beta=0.85$}]{\includegraphics[width=.245\textwidth]{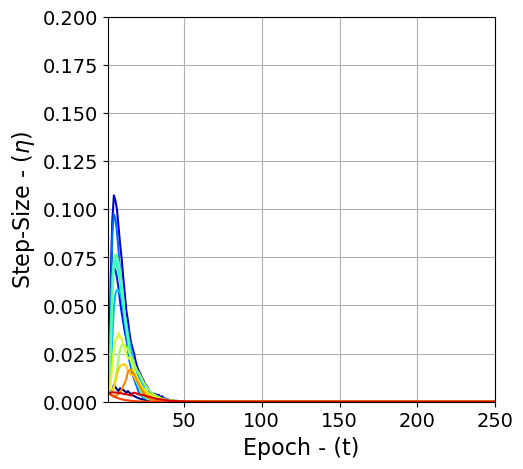}}
		\subfigure[\tiny{CIFAR10/VGG16/$\beta=0.90$}]{\includegraphics[width=.245\textwidth]{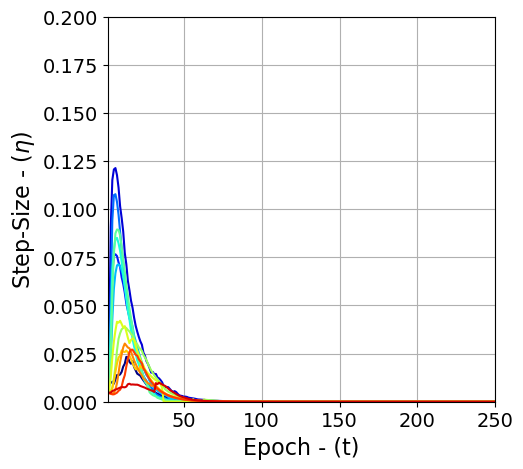}}
		\subfigure[\tiny{CIFAR10/VGG16/$\beta=0.95$}]{\includegraphics[width=.245\textwidth]{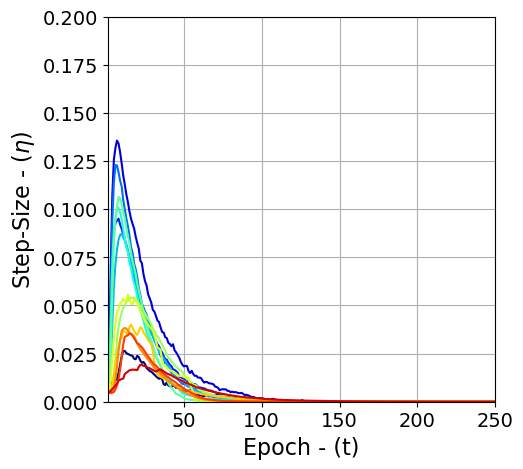}}
	}
	\centerline{
		\subfigure[\tiny{CIFAR10/ResNet34/$\beta=0.80$}]{\includegraphics[width=.245\textwidth]{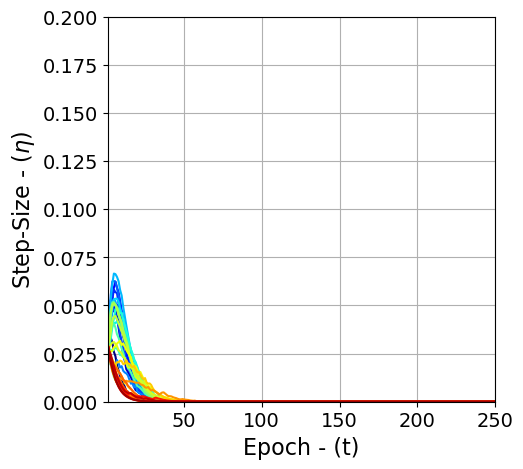}}
		\subfigure[\tiny{CIFAR10/ResNet34/$\beta=0.85$}]{\includegraphics[width=.245\textwidth]{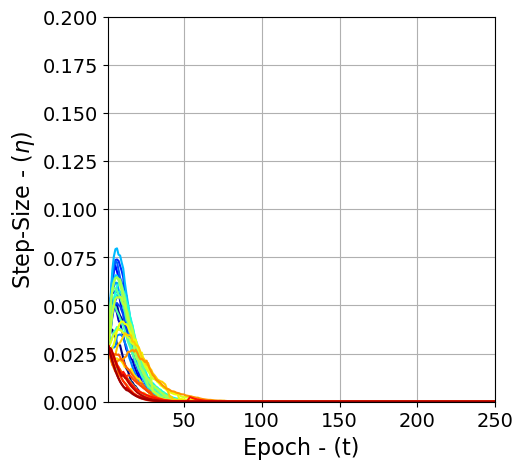}}
		\subfigure[\tiny{CIFAR10/ResNet34/$\beta=0.90$}]{\includegraphics[width=.245\textwidth]{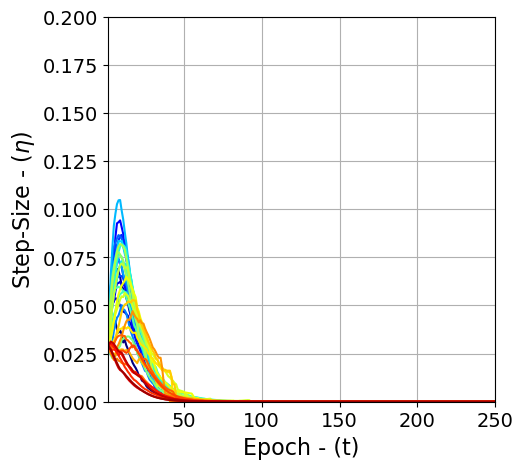}}
		\subfigure[\tiny{CIFAR10/ResNet34/$\beta=0.95$}]{\includegraphics[width=.245\textwidth]{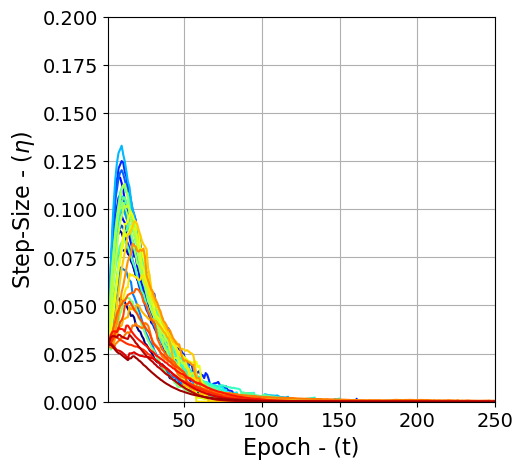}}
	}
	\centerline{
		\subfigure[\tiny{CIFAR100/VGG16/$\beta=0.80$}]{\includegraphics[width=.245\textwidth]{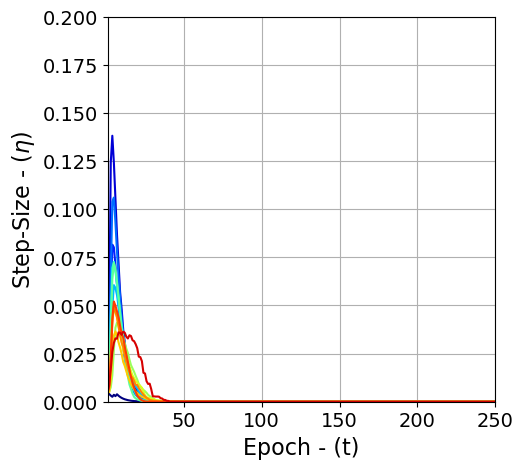}}
		\subfigure[\tiny{CIFAR100/VGG16/$\beta=0.85$}]{\includegraphics[width=.245\textwidth]{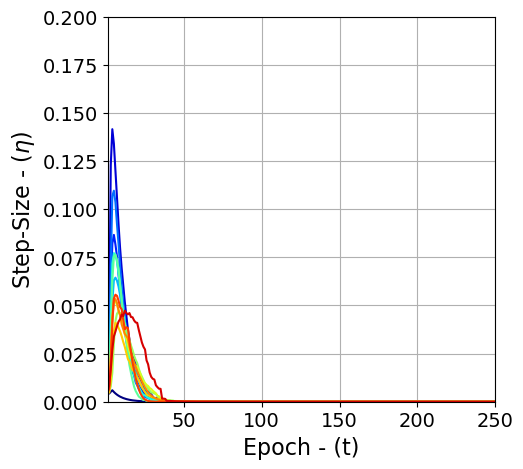}}
		\subfigure[\tiny{CIFAR100/VGG16/$\beta=0.90$}]{\includegraphics[width=.245\textwidth]{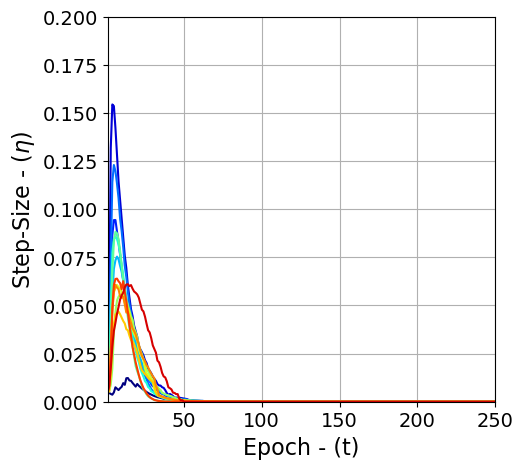}}
		\subfigure[\tiny{CIFAR100/VGG16/$\beta=0.95$}]{\includegraphics[width=.245\textwidth]{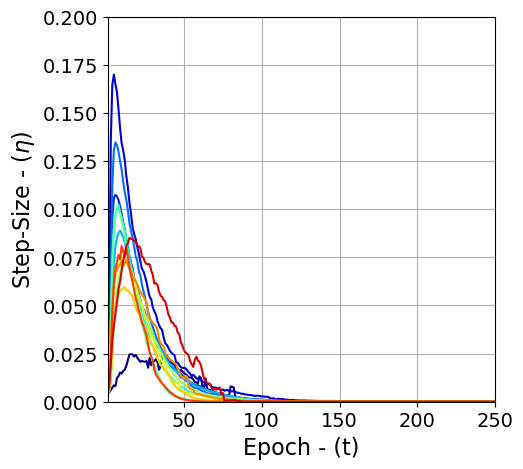}}
	}
	\centerline{
		\subfigure[\tiny{CIFAR100/ResNet34/$\beta=0.80$}]{\includegraphics[width=.245\textwidth]{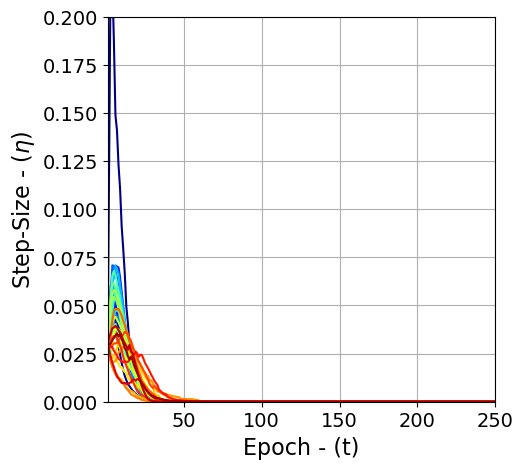}}
		\subfigure[\tiny{CIFAR100/ResNet34/$\beta=0.85$}]{\includegraphics[width=.245\textwidth]{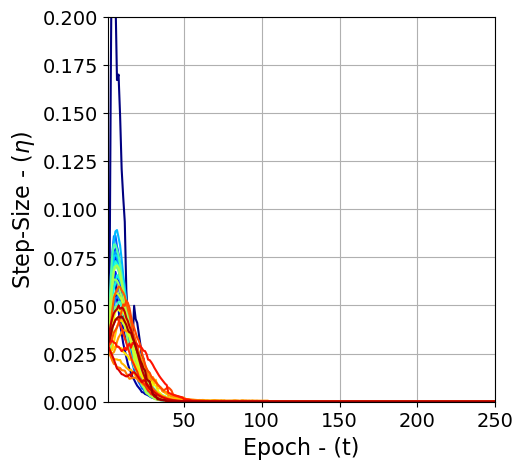}}
		\subfigure[\tiny{CIFAR100/ResNet34/$\beta=0.90$}]{\includegraphics[width=.245\textwidth]{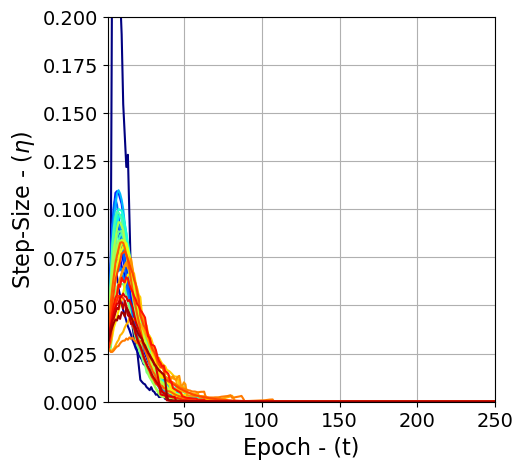}}
		\subfigure[\tiny{CIFAR100/ResNet34/$\beta=0.95$}]{\includegraphics[width=.245\textwidth]{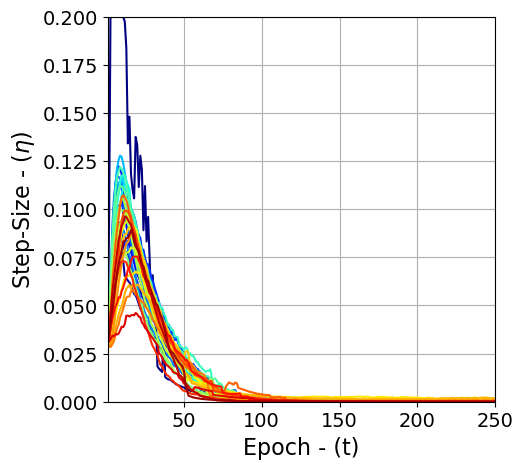}}
	}
	\caption{Ablative study of AdaS momentum rate $\beta$ versus mapping condition $\kappa$ over two different datasets (i.e. CIFAR10 and CIFAR100) and two CNNs (i.e. VGG16 and ResNet34). The transition in color shades from light to dark lines correspond to first to the end of convolution layers in each network.}
	\label{fig_ablative_study_AdaS_momentum_rate_vs_learning_rate}
\end{figure}

\begin{figure}[htp]
	\centerline{
		\subfigure[\tiny{VGG16, AdaGrad}]{\includegraphics[width=.16\textwidth]{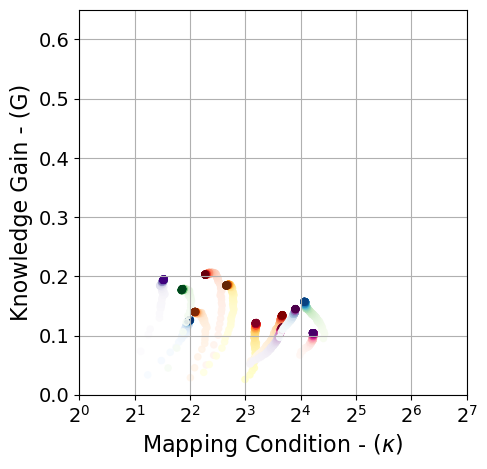}}
		\subfigure[\tiny{VGG16, RMSProp}]{\includegraphics[width=.16\textwidth]{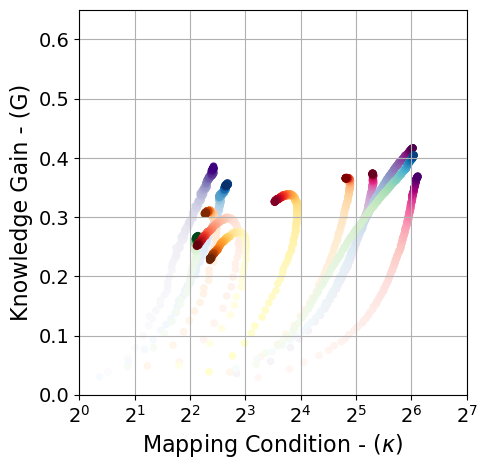}}
		\subfigure[\tiny{VGG16, AdaM}]{\includegraphics[width=.16\textwidth]{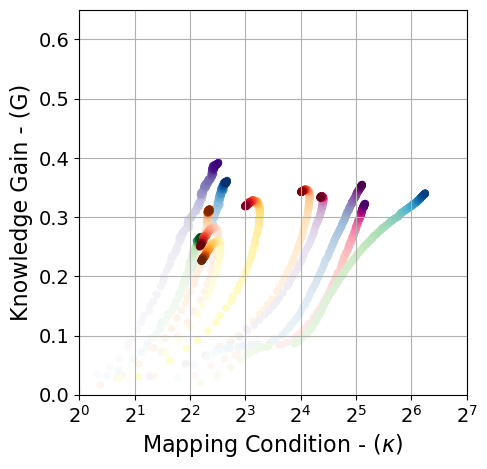}}
		\subfigure[\tiny{VGG16, AdaBound}]{\includegraphics[width=.16\textwidth]{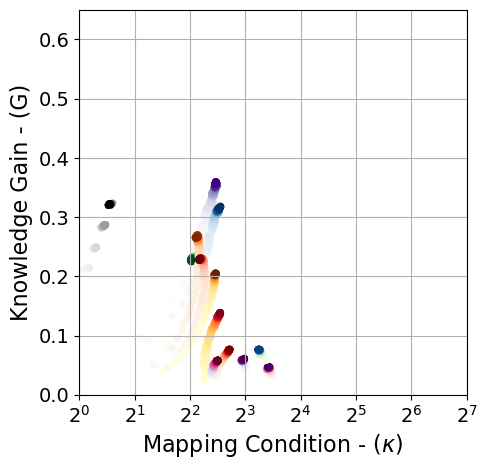}}
		\subfigure[\tiny{VGG16, StepLR}]{\includegraphics[width=.16\textwidth]{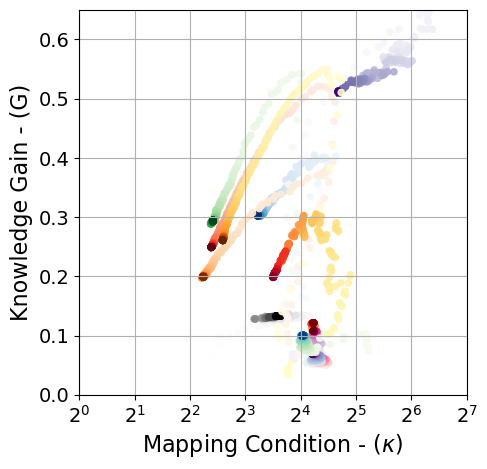}}
		\subfigure[\tiny{VGG16, OneCycleLR}]{\includegraphics[width=.16\textwidth]{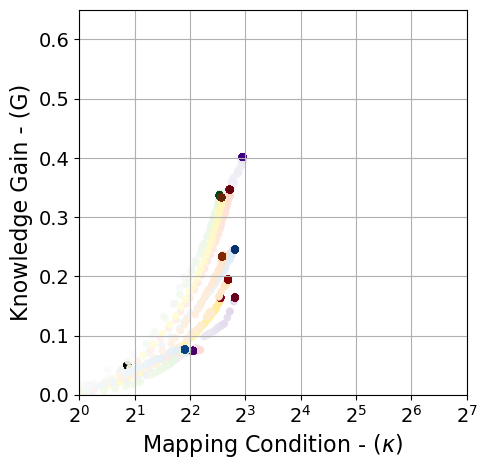}}
	}
	\centerline{
		\subfigure[\tiny{VGG16, AdaS-$\beta=0.80$}]{\includegraphics[width=.16\textwidth]{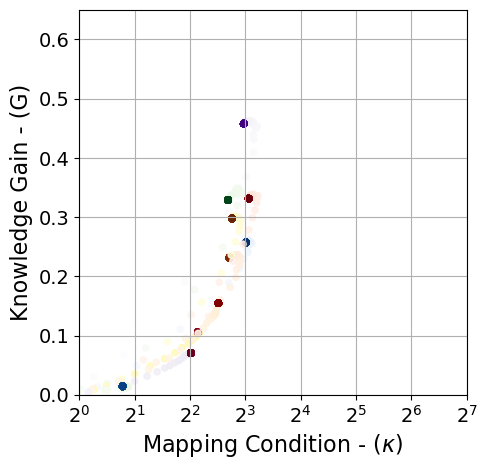}}
		\subfigure[\tiny{VGG16, AdaS-$\beta=0.85$}]{\includegraphics[width=.16\textwidth]{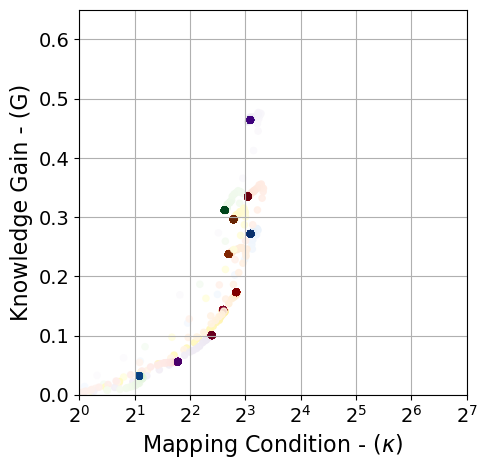}}
		\subfigure[\tiny{VGG16, AdaS-$\beta=0.90$}]{\includegraphics[width=.16\textwidth]{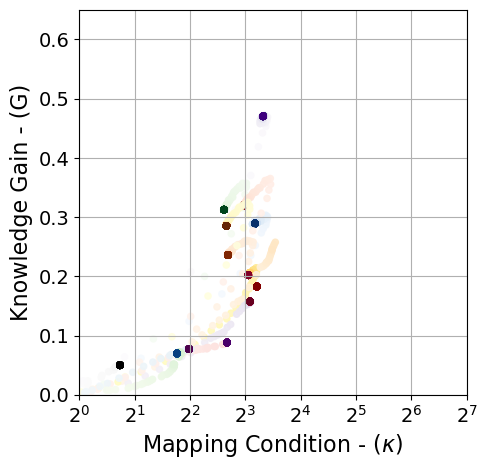}}
		\subfigure[\tiny{VGG16, AdaS-$\beta=0.925$}]{\includegraphics[width=.16\textwidth]{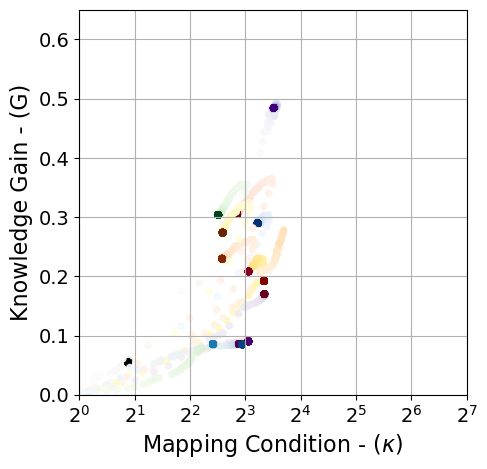}}
		\subfigure[\tiny{VGG16, AdaS-$\beta=0.95$}]{\includegraphics[width=.16\textwidth]{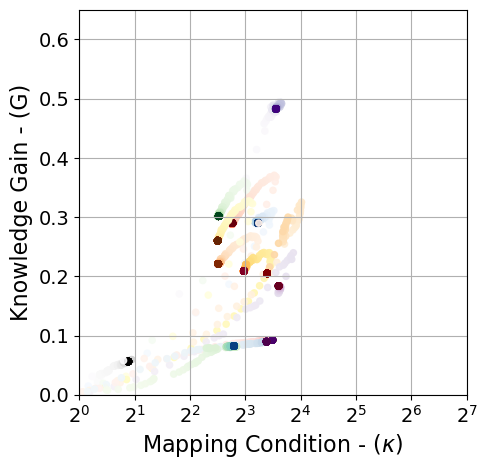}}
		\subfigure[\tiny{VGG16, AdaS-$\beta=0.975$}]{\includegraphics[width=.16\textwidth]{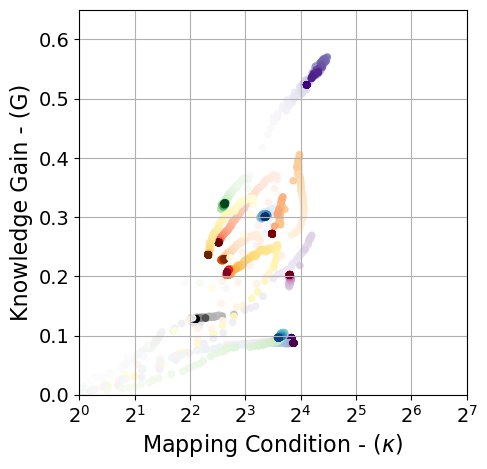}}
	}
	\centerline{
		\subfigure[\tiny{ResNet34, AdaGrad}]{\includegraphics[width=.16\textwidth]{knowledge_gain_vs_mapping_condition_CIFAR10_ResNet34_AdaGrad.png}}
		\subfigure[\tiny{ResNet34, RMSProp}]{\includegraphics[width=.16\textwidth]{knowledge_gain_vs_mapping_condition_CIFAR10_ResNet34_RMSProp.png}}
		\subfigure[\tiny{ResNet34, AdaM}]{\includegraphics[width=.16\textwidth]{knowledge_gain_vs_mapping_condition_CIFAR10_ResNet34_AdaM.png}}
		\subfigure[\tiny{ResNet34, AdaBound}]{\includegraphics[width=.16\textwidth]{knowledge_gain_vs_mapping_condition_CIFAR10_ResNet34_AdaBound.png}}
		\subfigure[\tiny{ResNet34, StepLR}]{\includegraphics[width=.16\textwidth]{knowledge_gain_vs_mapping_condition_CIFAR10_ResNet34_SGD_StepLR.png}}
		\subfigure[\tiny{ResNet34, OneCycleLR}]{\includegraphics[width=.16\textwidth]{knowledge_gain_vs_mapping_condition_CIFAR10_ResNet34_SGD_OneCycleLR.png}}
	}
	\centerline{
		\subfigure[\tiny{ResNet34, AdaS-$\beta=0.80$}]{\includegraphics[width=.16\textwidth]{knowledge_gain_vs_mapping_condition_CIFAR10_ResNet34_AdaS_beta_0800.png}}
		\subfigure[\tiny{ResNet34, AdaS-$\beta=0.85$}]{\includegraphics[width=.16\textwidth]{knowledge_gain_vs_mapping_condition_CIFAR10_ResNet34_AdaS_beta_0850.png}}
		\subfigure[\tiny{ResNet34, AdaS-$\beta=0.90$}]{\includegraphics[width=.16\textwidth]{knowledge_gain_vs_mapping_condition_CIFAR10_ResNet34_AdaS_beta_0900.png}}
		\subfigure[\tiny{ResNet34, AdaS-$\beta=0.925$}]{\includegraphics[width=.16\textwidth]{knowledge_gain_vs_mapping_condition_CIFAR10_ResNet34_AdaS_beta_0925.png}}
		\subfigure[\tiny{ResNet34, AdaS-$\beta=0.95$}]{\includegraphics[width=.16\textwidth]{knowledge_gain_vs_mapping_condition_CIFAR10_ResNet34_AdaS_beta_0950.png}}
		\subfigure[\tiny{ResNet34, AdaS-$\beta=0.975$}]{\includegraphics[width=.16\textwidth]{knowledge_gain_vs_mapping_condition_CIFAR10_ResNet34_AdaS_beta_0975.png}}
	}
	\caption{Evolution of knowledge gain versus mapping condition over iteration of epoch training for CIFAR10 dataset. Transition of colors shades correspond to different convolution blocks. The transparency of scatter plots corresponds to the convergence in epochs--the higher the transparency, the faster the convergence.}
	\label{fig_training_quality_all_methods_CIFAR10}
\end{figure}

\begin{figure}[htp]
	\centerline{
		\subfigure[\tiny{VGG16, AdaGrad}]{\includegraphics[width=.16\textwidth]{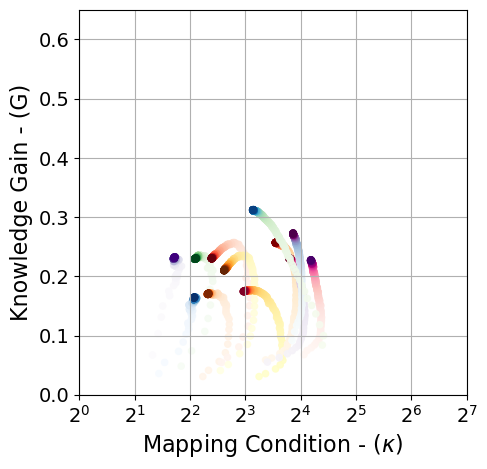}}
		\subfigure[\tiny{VGG16, RMSProp}]{\includegraphics[width=.16\textwidth]{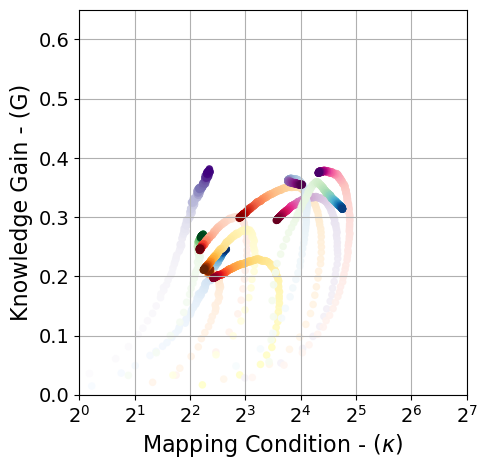}}
		\subfigure[\tiny{VGG16, AdaM}]{\includegraphics[width=.16\textwidth]{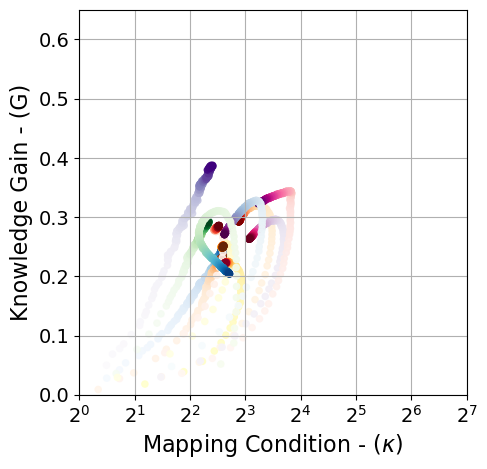}}
		\subfigure[\tiny{VGG16, AdaBound}]{\includegraphics[width=.16\textwidth]{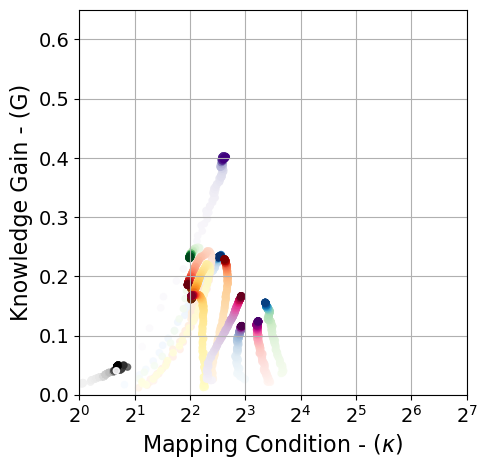}}
		\subfigure[\tiny{VGG16, StepLR}]{\includegraphics[width=.16\textwidth]{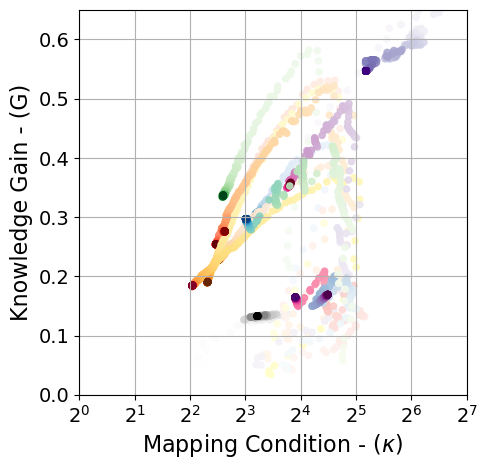}}
		\subfigure[\tiny{VGG16, OneCycleLR}]{\includegraphics[width=.16\textwidth]{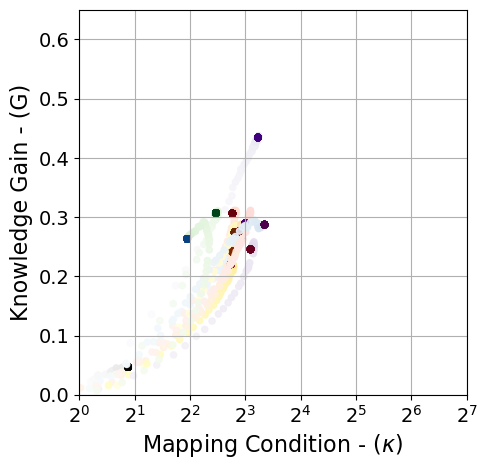}}
	}
	\centerline{
		\subfigure[\tiny{VGG16, AdaS-$\beta=0.80$}]{\includegraphics[width=.16\textwidth]{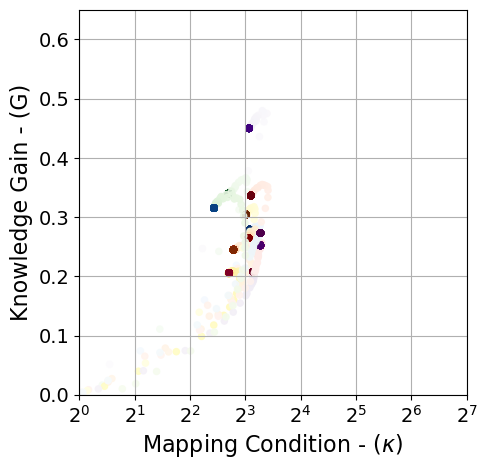}}
		\subfigure[\tiny{VGG16, AdaS-$\beta=0.85$}]{\includegraphics[width=.16\textwidth]{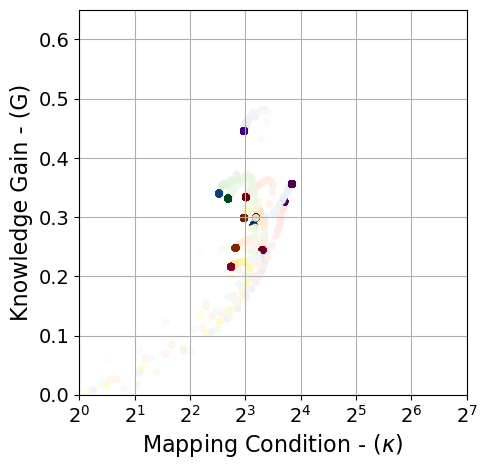}}
		\subfigure[\tiny{VGG16, AdaS-$\beta=0.90$}]{\includegraphics[width=.16\textwidth]{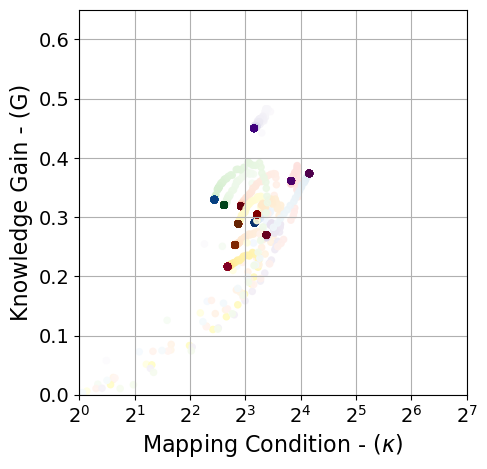}}
		\subfigure[\tiny{VGG16, AdaS-$\beta=0.925$}]{\includegraphics[width=.16\textwidth]{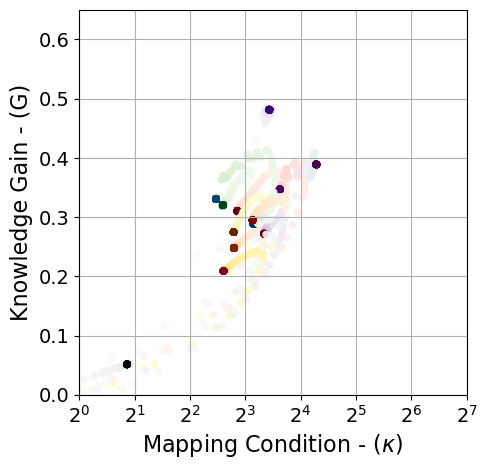}}
		\subfigure[\tiny{VGG16, AdaS-$\beta=0.95$}]{\includegraphics[width=.16\textwidth]{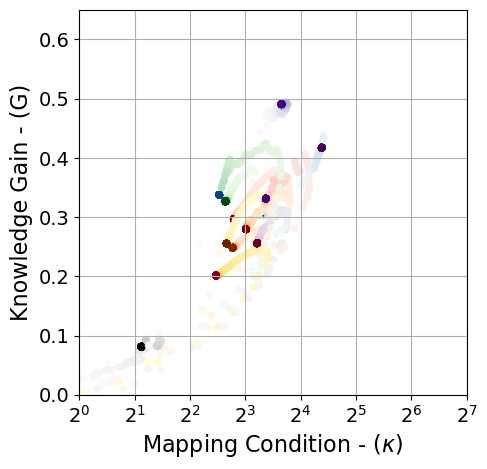}}
		\subfigure[\tiny{VGG16, AdaS-$\beta=0.975$}]{\includegraphics[width=.16\textwidth]{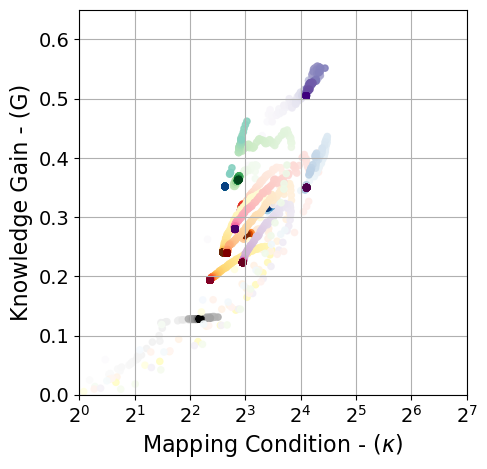}}
	}
	\centerline{
		\subfigure[\tiny{ResNet34, AdaGrad}]{\includegraphics[width=.16\textwidth]{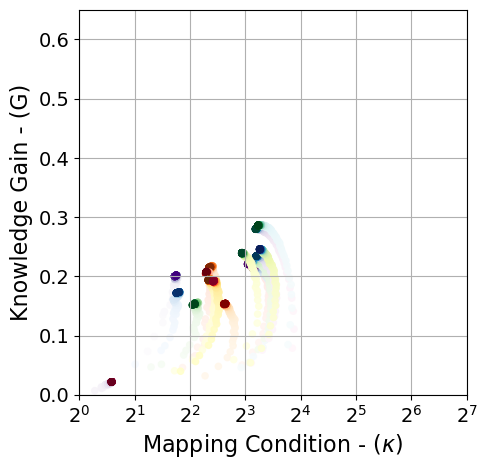}}
		\subfigure[\tiny{ResNet34, RMSProp}]{\includegraphics[width=.16\textwidth]{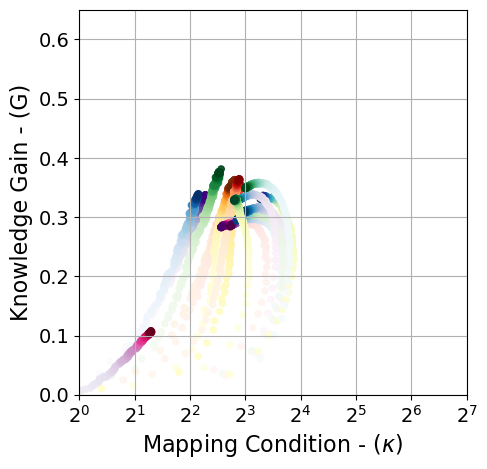}}
		\subfigure[\tiny{ResNet34, AdaM}]{\includegraphics[width=.16\textwidth]{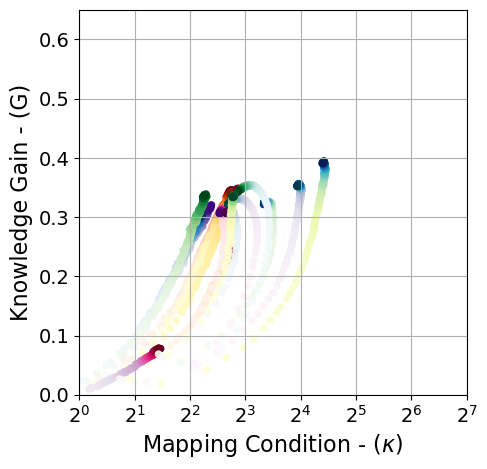}}
		\subfigure[\tiny{ResNet34, AdaBound}]{\includegraphics[width=.16\textwidth]{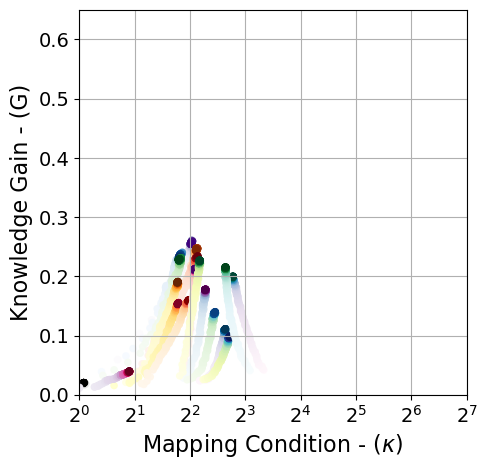}}
		\subfigure[\tiny{ResNet34, StepLR}]{\includegraphics[width=.16\textwidth]{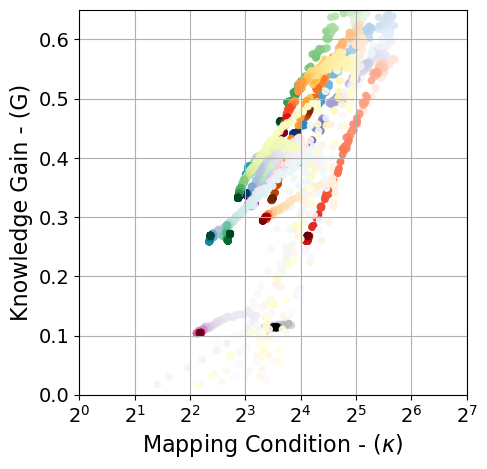}}
		\subfigure[\tiny{ResNet34, OneCycleLR}]{\includegraphics[width=.16\textwidth]{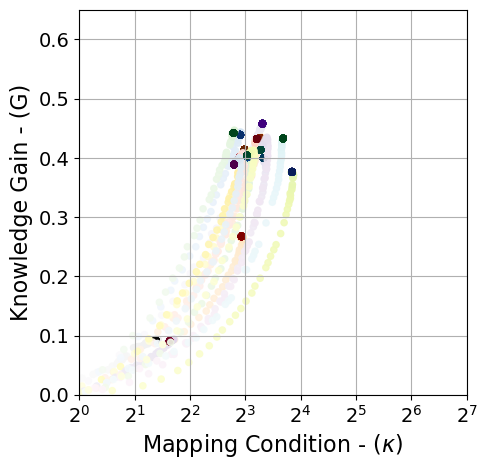}}
	}
	\centerline{
		\subfigure[\tiny{ResNet34, AdaS-$\beta=0.80$}]{\includegraphics[width=.16\textwidth]{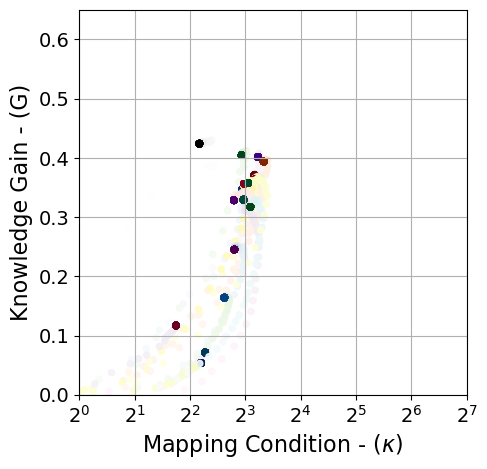}}
		\subfigure[\tiny{ResNet34, AdaS-$\beta=0.85$}]{\includegraphics[width=.16\textwidth]{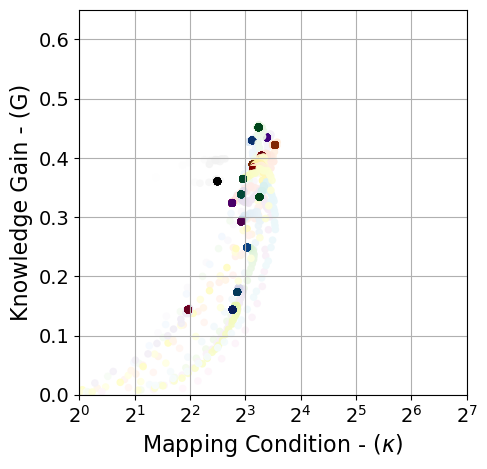}}
		\subfigure[\tiny{ResNet34, AdaS-$\beta=0.90$}]{\includegraphics[width=.16\textwidth]{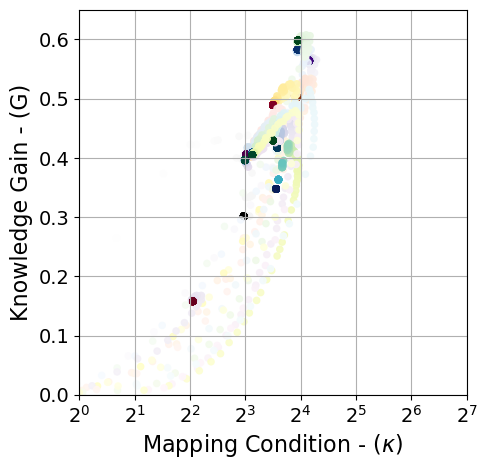}}
		\subfigure[\tiny{ResNet34, AdaS-$\beta=0.925$}]{\includegraphics[width=.16\textwidth]{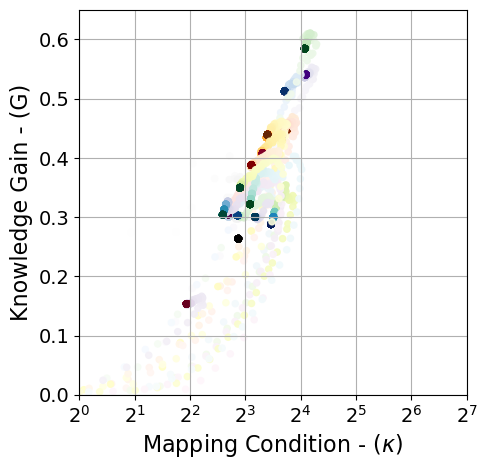}}
		\subfigure[\tiny{ResNet34, AdaS-$\beta=0.95$}]{\includegraphics[width=.16\textwidth]{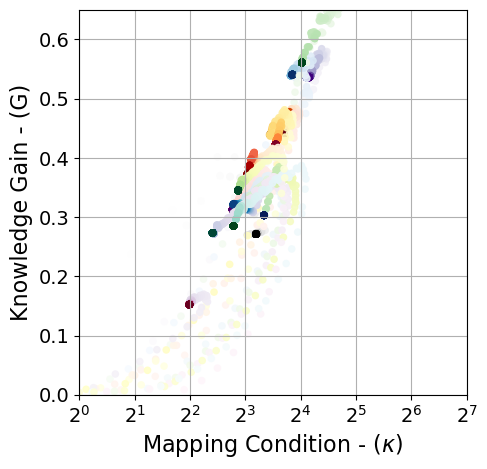}}
		\subfigure[\tiny{ResNet34, AdaS-$\beta=0.975$}]{\includegraphics[width=.16\textwidth]{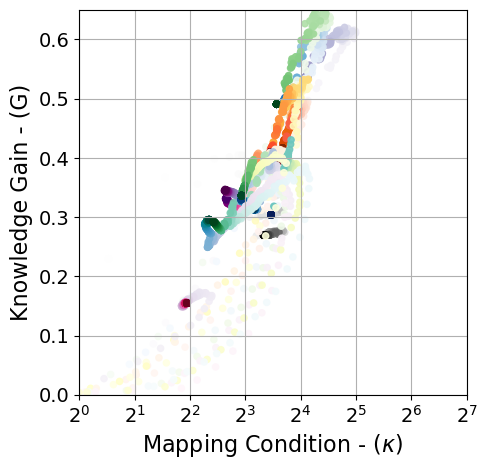}}
	}
	\caption{Evolution of knowledge gain versus mapping condition over iteration of epoch training for CIFAR100 dataset. Transition of colors shades correspond to different convolution blocks. The transparency of scatter plots corresponds to the convergence in epochs--the higher the transparecy, the faster the convergence.}
	\label{fig_training_quality_all_methods_CIFAR100}
\end{figure}

The ablative analysis of AdaS optimizer is studied here with respect to different parameter settings. Figure \ref{fig_ablative_study_dataset_vs_network} demonstrates the AdaS performance with respect to different range of gain-factor $\beta$. Figure \ref{fig_ablative_study_AdaS_momentum_rate_vs_knowledge_gain} demonstrates the knowledge gain of different dataset and network with respect to different gain-factor settings over successive epochs. Similarly, Figure \ref{fig_ablative_study_AdaS_momentum_rate_vs_rank_gain} also demonstrates the rank gain (aka the ratio of non-zero singular values of low-rank structure with respect to channel size) over successive epochs. Mapping conditions are shown in Figure \ref{fig_ablative_study_AdaS_momentum_rate_vs_mapping_condition} and Figure \ref{fig_ablative_study_AdaS_momentum_rate_vs_learning_rate} demonstrates the learning rate approximation through AdaS algorithm over successive epoch training. Evolution of knowledge gain versus mapping conditions are also shown in Figure \ref{fig_training_quality_all_methods_CIFAR10} and Figure \ref{fig_training_quality_all_methods_CIFAR100}.

\end{document}